\pdfoutput=1
\documentclass[11pt]{article}
\usepackage{acl} % убери [review] для camera-ready

\usepackage[T1]{fontenc}
\usepackage[utf8]{inputenc}
\usepackage{times}
\usepackage{latexsym}
\usepackage{microtype}

\usepackage{amsmath}
\usepackage{amsfonts}

\usepackage{graphicx}
\usepackage{float}
\usepackage{wrapfig}
\usepackage{subcaption}
\usepackage{caption}

\usepackage{booktabs}
\usepackage{tabularx}
\usepackage{tabulary}
\usepackage{multirow}
\usepackage{makecell}
\usepackage{array}
\usepackage{arydshln}
\usepackage{colortbl}
\usepackage{xcolor}

\definecolor{soft_teal}{HTML}{2FBEAD}
\definecolor{soft_tan}{HTML}{E8D5C4}
\definecolor{custombg}{HTML}{DFF9FB}
\definecolor{codegreen}{rgb}{0,0.6,0}
\definecolor{codegray}{rgb}{0.5,0.5,0.5}
\definecolor{codepurple}{rgb}{0.58,0,0.82}

\usepackage{algorithm}
\usepackage{algorithmic}
\usepackage{listings}
\usepackage{pifont}
\usepackage{mdframed}
\usepackage{tcolorbox}
\usepackage{fancyvrb}
\DefineVerbatimEnvironment{MyVerbatim}{Verbatim}{
  fontsize=\scriptsize,
  breaklines=true
}

\usepackage{ifthen}
\newboolean{showcomments}
\setboolean{showcomments}{true}

\ifthenelse{\boolean{showcomments}}{
  \newcommand{\nb}[3]{
    {\color{#2}\small\fbox{\bfseries\sffamily\scriptsize#1}}
    {\color{#2}\sffamily\small$\triangleright$~\textit{#3}~$\triangleleft$}
  }
}{
  \newcommand{\nb}[3]{}
}

\usepackage{hyperref}
\usepackage{url}

\title{Self-Guided Plan Extraction for Instruction-Following Tasks with Goal-Conditional Reinforcement Learning}

\author{
Zoya Volovikova$^{1,2}$ \quad
Nikita Sorokin$^{1}$ \quad
Dmitriy Lukashevskiy$^{2}$ \\
\bf Aleksandr Panov$^{1,2}$ \quad
Alexey Skrynnik$^{1,2}$ \\
\\
$^1$AXXX, $^2$MIRAI \\
}
  
\begin{document}

\maketitle

\begin{abstract}
We introduce SuperIgor, a framework for instruction-following tasks. Unlike prior methods that rely on predefined subtasks, SuperIgor enables a language model to generate and refine high-level plans through a self-learning mechanism, reducing the need for manual dataset annotation. Our approach involves iterative co-training: an RL agent is trained to follow the generated plans, while the language model adapts and modifies these plans based on RL feedback and preferences. This creates a feedback loop where both the agent and the planner improve jointly. We validate our framework in environments with rich dynamics and stochasticity. Results show that SuperIgor agents adhere to instructions more strictly than baseline methods, while also demonstrating strong generalization to previously unseen instructions.
\end{abstract}

\section{Introduction}

The instruction-following task~\citep{shridhar2020alfred, chevalier2018babyai, zhong2021silg} studies how an AI agent can achieve a goal specified through a natural-language instruction. Prior work in this area is commonly divided into two paradigms: learning from demonstrations and learning from experience. Demonstration-based approaches rely on expert trajectories (e.g., VPT~\citep{baker2022video}, STEVE-1~\citep{lifshitz2023steve}, Jarvis-VLA~\citep{li2025jarvis}) and exhibit strong generalization to unseen tasks and environments~\citep{fan2022minedojo, zhou2024minedreamer, gray2019craftassistframeworkdialogueenabledinteractive}. However, their scalability is fundamentally constrained by the high cost of collecting large quantities of high-quality demonstrations, as highlighted in SIMA~2~\citep{bolton2025sima}.

\begin{figure}[ht!]
    \centering
    \includegraphics[width=1\linewidth]{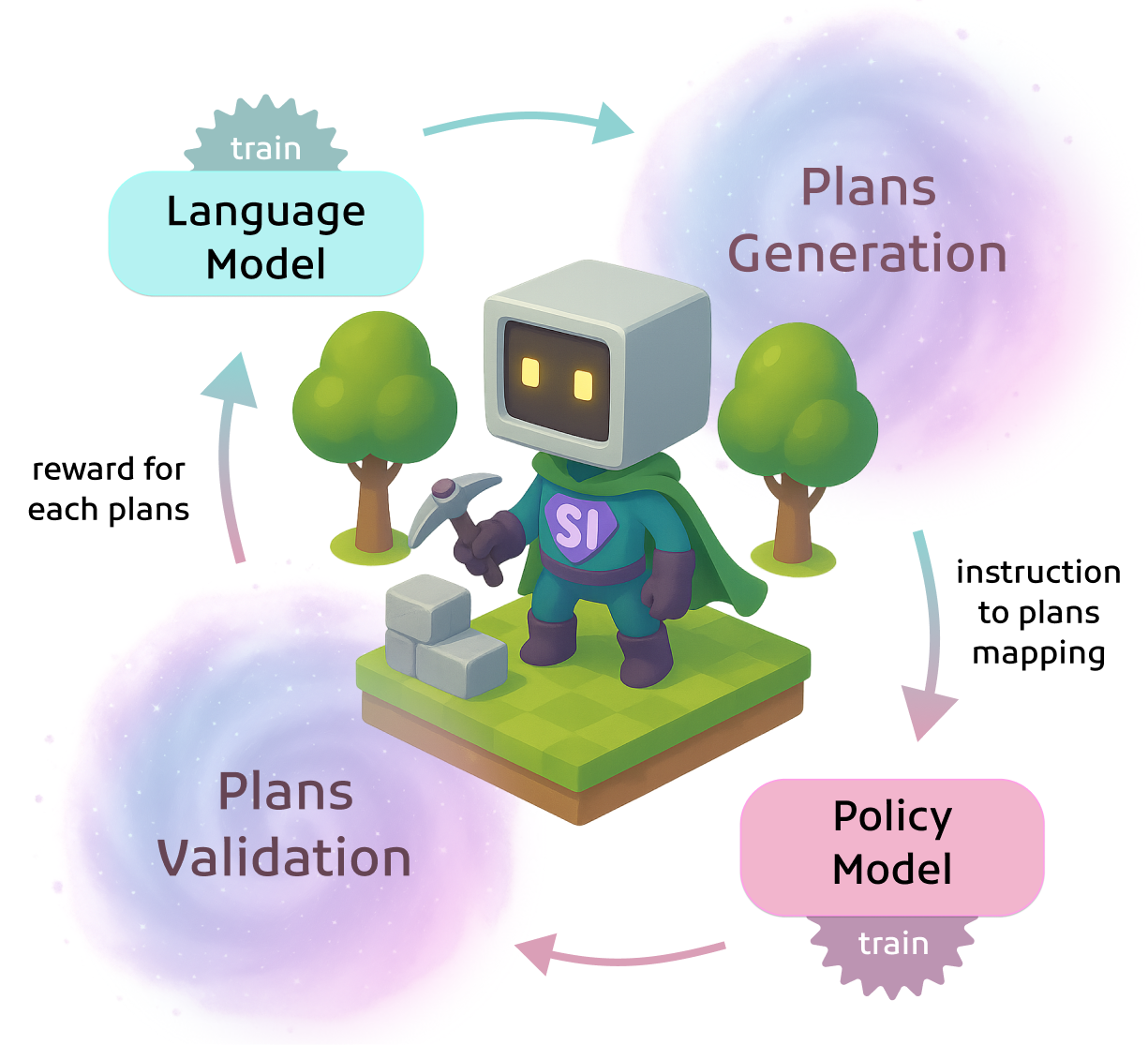}
    \caption{Conceptual diagram of the \textbf{SuperIgor} framework designed for Instruction Following}
    \label{fig:visual_abstract}
    \vspace{-10px}
\end{figure}

In contrast, experience-based methods, typically framed as reinforcement learning, learn directly from the agent’s own interactions with the environment without expert supervision~\citep{hill2020grounded, mathur2025adapting}. While conceptually more scalable, these methods face substantially greater challenges, most notably the difficulty of grounding natural-language instructions in long-horizon behavior under sparse and delayed rewards~\citep{chevalier2018babyai, hanjie2021grounding, zhong2019rtfm}. To make progress despite these challenges, most experience-based instruction-following methods have been studied in controlled and simplified environments, such as grid-world or cell-based domains, where observations are often symbolic rather than pixel-based, instructions are procedurally generated, and linguistic diversity is limited~\citep{volovikova2025craftext}.

A natural way to strengthen experience-based instruction-following agents is to incorporate high-level planning through language models, which helps interpret complex and underspecified tasks. By leveraging world knowledge encoded in large language models, such planning can reduce the exploration space and accelerate learning~\citep{jansen2020visually, zhou2407autonomous}. Existing planning-based approaches typically adopt a common design choice: planning is performed over a predefined and finite set of subgoals whose completion can be explicitly verified by the environment. This assumption enables generated subgoals to be grounded in an existing skill library via heuristics or similarity-based matching~\citep{logeswaran2022few}, provides dense intermediate rewards, and allows automatic skill switching during execution, thereby bypassing the need to learn low-level policies from scratch~\citep{ahn2022can}. Furthermore, many planning-based systems rely on very large language models (100B+ parameters), which further limits their practicality and scalability.

These limitations raise an important question: how can we learn a low-level policy for instruction following in environments without predefined low-level skills? We address this challenge by proposing SuperIgor, a reinforcement-learning-based framework that integrates high-level planning through large language models. SuperIgor adopts a self-learning strategy that allows the agent to iteratively refine its plans based on its own experience, and to learn effectively from sparse and delayed instruction-level rewards provided only upon successful task completion. We further demonstrate that SuperIgor operates effectively in CrafText, a dynamic, partially observable, and open-ended environment, highlighting its applicability beyond simplified instruction-following benchmarks.

To conclude, our contributions are as follows:

\begin{itemize}
    \item We propose a new self-supervised training paradigm for the instruction-following task, where high-level plans are generated and refined through interaction between a language model and a reinforcement learning agent—without requiring any manually annotated datasets.

    \item We introduce a special curriculum to train an RL agent to accurately follow the plan despite sparse reward conditions.

    \item We implement our approach in the CrafText benchmark and achieve state-of-the-art performance on out-of-distribution tasks, demonstrating the robustness and flexibility of our framework in dynamic and partially observable environments.  The dataset and code for SuperIgor are publicly available\footnote{\url{https://sites.google.com/view/super-igor}}.  
\end{itemize}

\section{Related Work}

\textbf{Instruction Following with RL}. One of the most common strategies for training agents with RL on instruction-following tasks is to jointly encode the instruction and the agent’s observations, enabling alignment between linguistic and perceptual modalities. A prominent line of work relies on shared representation models such as CLIP~\citep{yao2022detclip}, or feature modulation techniques like FiLM layers~\citep{perez2018film, chevalier2018babyai, zhong2019rtfm}, to project language information into visual or state representations~\citep{paischer2023semantic, hanjie2021grounding, zhong2021silg}. Alternatively, transformer-based architectures process multimodal inputs jointly to improve instruction understanding and execution. This includes embodied language models such as EmBERT~\citep{suglia2108embodied} as well as Vision-and-Language Navigation frameworks~\citep{savva2019habitat}. Another direction explores model-based reinforcement learning, where agents learn structured policies conditioned on textual goals; Dynalang~\citep{Dynalang} is a representative example of this approach, emphasizing learning world dynamics alongside goal-conditioned behavior.

\textbf{Instruction Following and Planning}. Recent work has demonstrated that large language models, when fine-tuned on suitable datasets, are capable of generating detailed, coherent action plans for agents based on textual instructions~\citep{jansen2020visually, zhou2407autonomous}. Building on this ability, subsequent approaches have shown that planning performance can be further improved by incorporating feedback from the environment~\citep{wang2023describe, huang2022language, volovikova2024instruction}. For example SayCan~\citep{ahn2022can} augments planning with affordance-based evaluation via offline reinforcement learning, while~\citep{tan2024true} leverages policy optimization and probability normalization to enhance learning through interaction. Beyond improving general plan quality, environment feedback can also enable personalized planning; for example, \cite{han2024llm} introduced Reinforced Self-Training to iteratively align agents' behavior with user preferences in object rearrangement tasks. Alternatively, \citet{logeswaran2022few} proposed a different strategy by avoiding language model fine-tuning altogether, instead generating multiple candidate plans with a frozen model and ranking them using mutual information and a learned feasibility model.

\section{Problem Statement}
\label{sec:rl_plan_setup}
The environment is formalized as a goal-based Partially Observable Markov Decision Process (POMDP), defined by the tuple $(\mathcal{S}, \mathcal{A}, \mathcal{O}, \mathcal{T}, \mathcal{R}, \mathcal{G}, \gamma)$.  
The agent receives a natural language instruction $I$ and must achieve the corresponding latent goal $g \in \mathcal{G}$. Each observation $o \in \mathcal{O}$ contains partial information about both the environment and the instruction $I$.  
The agent learns a grounding function $f_g(I)$ to infer the latent goal $g = f_g(I)$.

The policy $ \pi(a \mid o) $ selects actions based on observations to maximize the expected cumulative reward: $$ \pi^* = \arg\max_{\pi} \mathbb{E}_\pi\left[\sum_{t=0}^{T} \gamma^t R(s_t, a_t, g) \,\middle|\, o_0\right].$$ The environment involves stochastic transitions $\mathcal{T}(s' \mid s, a)$ and partial observability, requiring the agent to infer goals and act effectively under uncertainty.  

We extend this setup by \textbf{introducing plans}. In the planning-augmented formulation, the agent does not receive the instruction $I$ directly.  Instead, it is provided with a plan $p = (p_1, p_2, \ldots, p_n)$ derived from $I$, where each step $p_i$ corresponds to an intermediate subgoal $g_i = f_g(p_i)$. At each timestep, the agent observes the environment together with the current plan step $p_{\phi(t)}$.  
The optimization objective becomes: $$
\pi^* = \arg\max_{\pi} \mathbb{E}_\pi\left[\sum_{t=0}^{T} \gamma^t R\left(s_t, a_t, g_{\phi(t)}\right) \,\middle|\, o_0\right], $$ where $g_{\phi(t)}$ is the subgoal associated with the active plan step.  

In contrast to settings with predefined subtasks and explicit intermediate rewards, our formulation introduces two key challenges:  

\begin{enumerate}
    \item \textbf{Subtask alignment under sparse rewards.}  
    The agent must discover how its behavior aligns with intermediate subgoals despite only receiving sparse, delayed feedback upon completing the full instruction.  
    This exacerbates the credit assignment problem.  

    \item \textbf{Extended action space.}  
    The agent must also decide when to terminate the current subtask.  
    This requires augmenting the action space with control operations (e.g., a \emph{DONE} action), which increases both exploration complexity and the difficulty of learning effective switching strategies.  
\end{enumerate}

\section{Super Igor}

\begin{figure*}[!ht]
    \centering
    \includegraphics[width=0.9\linewidth]{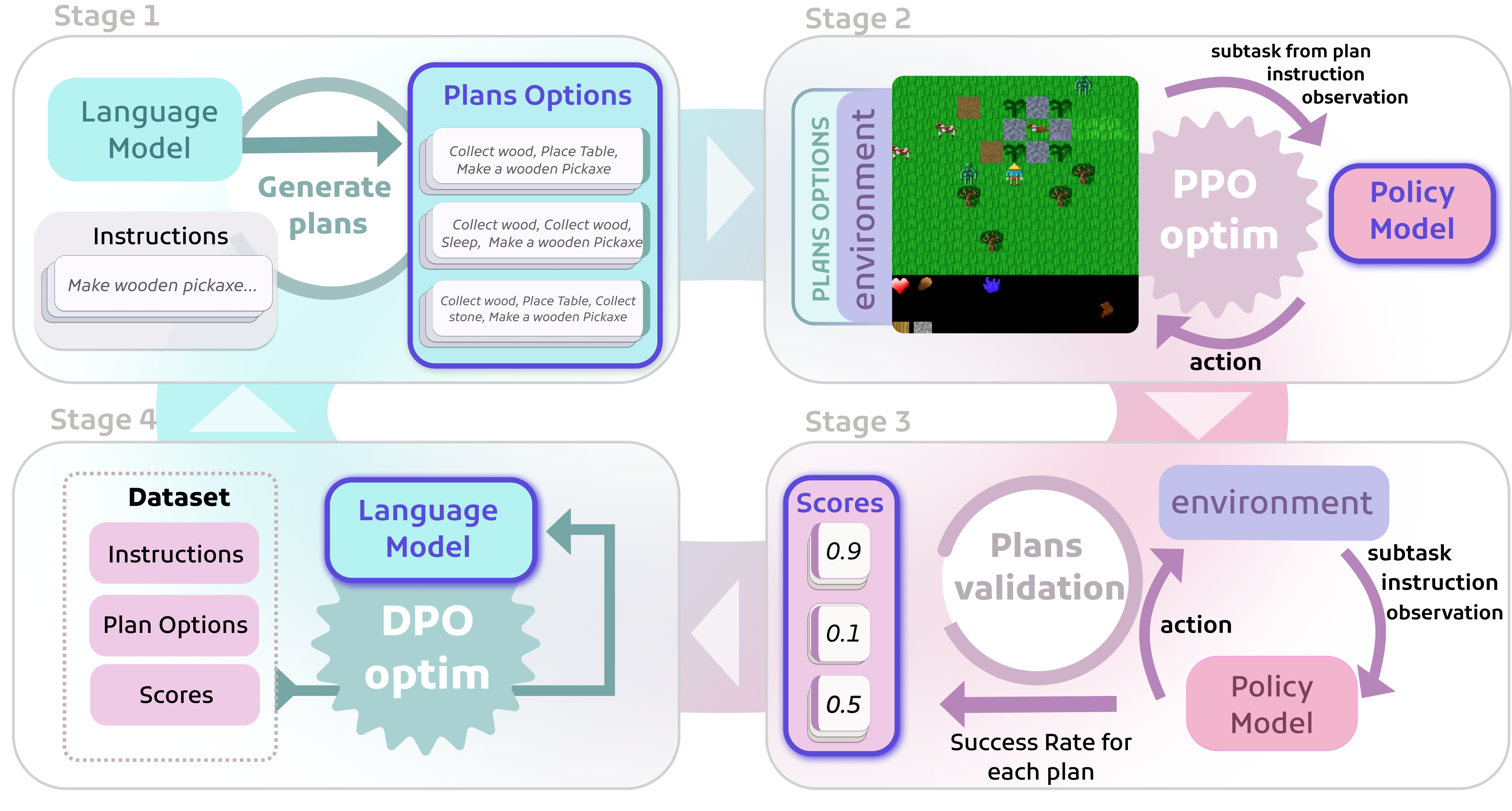}
    \caption{Super Igor Pipeline: The pipeline consists of four stages: (1) a language model generates multiple plan options for a given instruction; (2) a policy model is trained via PPO to execute each plan in the environment; (3) each plan is validated by measuring its execution success rate; (4) the language model is optimized using Direct Preference Optimization (DPO)\citep{rafailov2023direct} based on plan performance scores. This iterative loop refines both plan generation and execution.}
    \label{fig:si_pipe}
\vspace{-15px}
\end{figure*}
 Super Igor framework proposes a method for jointly training a large language model and a reinforcement learning  agent to solve instruction-following tasks. The LLM is responsible for transforming natural language instructions into structured plans, i.e. sequences of subtasks. The RL agent learns to execute these plans in the environment by interacting with it and maximizing delayed rewards.

The training process proceeds through the following stages:
\begin{enumerate}
\item \textbf{Plan Generation}~(\ref{subsection:SI_subgoals_generation}): The LLM extracts possible subtasks from instructions and generates multiple candidate plans in natural language during the initial cycle (Cycle~1). In subsequent cycles (Cycle~2–N), the candidate pool is iteratively refined by filtering and re-prioritization, based on how well the plans align with the RL agent’s performance.
\item \textbf{Policy Learning}~(\ref{subsection:SI_policy_learning}): The RL agent is trained to execute the selected plans in the environment.
\item \textbf{Plan Validation}~(\ref{subsection:SI_plan_validation}): The quality of candidate plans is evaluated according to the RL agent’s success rate and execution trajectories.
\item \textbf{LLM Fine-Tuning}~(\ref{subsection:SI_llm_finetuning}): The language model is fine-tuned with feedback derived from validation, aligning its scoring of plans with the agent’s actual performance.
\end{enumerate}

\subsection{Plan Generation}
\label{subsection:SI_subgoals_generation}

In our approach, we first generate all possible plans for the training set in zero-shot mode during the initial cycle. In subsequent cycles, we progressively reduce the set of candidate plans by filtering out those that perform poorly for the agent. Concretely, the initial cycle produces the complete pool of plans, while later cycles re-prioritize them using the LLM’s negative log-likelihood (NLL) score. Importantly, we leverage the agent’s performance feedback as a preference signal to fine-tune the LLM with DPO, so that the model learns to align its scoring with the agent’s actual success in executing the plans.

\textbf{Training plans generation (Cycle 1). }

% ----MORE INFORMAL---
Since the language model used for plan generation may not fully capture the exact dependencies and interaction rules of the target environment, we propose a structured procedure that separates the identification of goals from the reasoning about prerequisite constraints. The method unfolds in four steps. 

First, we build a \textit{subtask base} by extracting and canonicalizing possible subtasks from the instruction dataset, creating a unified vocabulary that reduces synonymy and ensures consistency. Each subtask is expressed in natural language, but in a strict normalized format that allows passing them one-by-one to the policy without ambiguity. 

Second, the model generates a \textit{goal-level plan}, producing for each instruction a single conceptual representation of its intended outcome, expressed in terms of the established subtask base. This step abstracts away from concrete execution details and captures only the high-level intent.

Third, we induce a \textit{ subtask ontology} that encodes the model’s hypotheses about prerequisite relations, i.e., which subtasks must be completed before others can be attempted. This provides a structured view of dependencies across the subtask base.

Finally, we perform \textit{plan expansion}, where the single conceptual plan is unfolded into multiple detailed plans, with their number corresponding to the hypotheses proposed by the model. The ontology ensures that these expanded variants remain consistent with prerequisite relations and avoid contradictions.

This approach provides two key benefits. First, it improves plan consistency by constructing plans from a shared set of subtasks and their relations, rather than from independent and potentially contradictory structures. Second, it supports partial normalization, since the model, when processing new instructions, tends to reuse previously identified subtasks, thereby reducing the proliferation of synonymous formulations. The details of the method and pseudocode are provided in Appendix \ref{app:zoe}, and the prompts are presented in Appendix \ref{app:zoe_prompts}.

\textbf{Plans re-prioretizing for RL-agent (Cycles 2-N).} After obtaining the initial feedback on agent performance for the generated plans and applying LLM fine-tuning (Subsection \ref{subsection:SI_llm_finetuning}), subsequent cycles focus on re-prioritizing the candidate set. In each cycle, plans are rescored using the language model’s negative log-likelihood (NLL), which reflects how natural or plausible a plan is according to the model. Plans are then ranked by this score, and only the top-performing subset is retained for further training. As cycles progress, this iterative filtering process gradually narrows the candidate space, aligning the remaining plans both with the agent’s empirical success and with the model’s learned preferences.

\subsection{Policy Learning}
\label{subsection:SI_policy_learning}

After the plans have been generated, we train a reinforcement learning agent using the stepwise plan observation setting (Subsection ~\ref{sec:rl_plan_setup}). At each timestep, the agent observes the environment and receives an embedding of the current plan step. It must learn to align actions with plan steps based on a delayed reward signal provided only upon successful completion of the entire plan. We use the PPO algorithm to train the policy.

\begin{algorithm}[H]
\footnotesize
\caption{Skill Curriculum Learning}
\label{alg:skill_curriculum}
\begin{algorithmic}[1]
\REQUIRE Set of all plans $\mathcal{P}$, success-rate threshold $\tau$

\STATE Initialize mastered skills $\mathcal{M} \gets \emptyset$
\STATE Initialize PPO agent $\pi_\theta$
\STATE Initialize active plans
\[
\mathcal{S} \gets \{p \in \mathcal{P} \mid p \text{ contains exactly one skill}\}
\]

\WHILE{training not converged}
    \STATE Train $\pi_\theta$ on active plans $\mathcal{S}$ and collect rollouts
    \STATE For each skill $s$, compute success rate:
    \[
    SR(s) = \frac{\text{\# Successful episodes containing } s}
                 {\text{\# Total episodes containing } s}
    \]
    \IF{$SR(s) \ge \tau$}
        \STATE Add to mastered skills $\mathcal{M} \gets \mathcal{M} \cup \{s\}$
    \ENDIF

    \STATE Update plans
    \[
    \mathcal{S} \gets \{p \in \mathcal{P} \mid p \text{ has at most one unmastered skill}\}
    \]
\ENDWHILE

\STATE \textbf{return} $\pi_\theta, \mathcal{M}$
\end{algorithmic}
\end{algorithm}

\vspace{-10px}

To address the sparse reward problem in training, we introduce \textbf{Skill Curriculum Learning}. The core principle is to create a dynamic curriculum that begins with the simplest single-subtask tasks, allowing the agent to learn foundational behaviors under a relatively dense reward signal.

As the agent trains, we monitor its Success Rate (SR) for each subtask. Once a subtask's SR surpasses a predefined threshold $\tau$, it is marked as "mastered." This mastery triggers an update to the curriculum: the set of active training plans is expanded to include any plan composed of already mastered subtasks and, at most, one new, unmastered subtask. This incremental expansion, detailed in Algorithm~\ref{alg:skill_curriculum}, ensures a smooth learning gradient and prevents the agent from being overwhelmed.

\subsection{Plan Validation}
\label{subsection:SI_plan_validation}

To evaluate each proposed plan, we run the RL agent multiple times using that plan as input. Because the environment is highly stochastic, a single rollout is insufficient; instead, we aggregate metrics such as average success rate  to obtain a reliable estimate of plan effectiveness.

\subsection{LLM Fine-Tuning}
\label{subsection:SI_llm_finetuning}

In the first cycle, we warm-start the language model with supervised fine-tuning (SFT) to reproduce the plans generated in the zero-shot stage (Section~\ref{subsection:SI_subgoals_generation}), aligning it with the plan distribution of the target environment.

In subsequent cycles, we incorporate plan-level quality signals obtained from execution and validation. These signals are used to construct preference pairs of higher- and lower-scoring plans, which are then used for DPO fine-tuning. This allows the model to internalize the agent’s feedback and gradually improve plan generation.

In our framework, DPO acts as a lightweight plan-selection bias rather than a precise credit assignment mechanism. It increases the probability of plan structures that the RL agent can learn from early, implicitly forming an automated curriculum, while deprioritizing plans that yield little initial progress without explicitly labeling them as incorrect.

\section{Experiments}

In this section, we describe the experiments conducted to answer the following research questions (RQ): %\input{iclr2026_conference}

\textbf{RQ1. (Effectiveness and Generalization of Auto-Generated Plans)}: How well can the SuperIgor agent learn to follow instructions by leveraging LLM-generated plans, and how well does this learned behavior generalize to new instructions?  We measure effectiveness as the agent's final success rate on training tasks (Atomic and Combo splits). We measure generalization using final success rates on two test sets: Paraphrases (same goals, new wording) and New Objects (new goal combinations).

\textbf{RQ2. (Policy Training under Sparse Feedback)}: How well can the SuperIgor policy model be trained to follow plans under sparse feedback?
The primary metric for this is the final SR on the training tasks. 

\textbf{RQ3. (Agent Effectiveness with Iterative SuperIgor Cycles)}: How does the agent's performance evolve over multiple iterations of the SuperIgor planning-training cycle?

\subsection{Environment} 
We conduct our experiments on the CrafText benchmark~\citep{volovikova2025craftext}, which provides a unified testbed for evaluating instruction-following agents in a multimodal, dynamic, and partially observable open-ended environment. Leveraging the modular design of CrafText, which allows for the procedural generation of custom tasks, we construct a specialized dataset named \textbf{FOCUS}.

The FOCUS dataset is built upon CrafText's Achievement scenarios but is specifically engineered to rigorously test instruction adherence under strict constraints. It contains over 900 instructions with a vocabulary of more than 1,500 unique words. The training set is composed of two instruction types: \textbf{Atomic}, which specify a single, indivisible goal (e.g., ``craft a furnace''), and \textbf{Combo}, which combine multiple atomic goals into a sequence of actions (e.g., ``craft a furnace and then collect wood'').

A critical challenge in this domain is that task composition often involves overlapping subtasks. For example, crafting a furnace first requires making a wooden pickaxe and collecting stone — steps that are also required for other goals. Consequently, agents may learn generic subroutines that maximize reward without truly grounding the linguistic instruction. To address this, FOCUS enforces a \textit{strict evaluation protocol}: an instruction is considered successful only if all its specified goals are completed precisely as requested, with no extraneous achievements triggered. This prevents the agent from exploiting broad, non-specific strategies.

To evaluate generalization, FOCUS employs two distinct test sets. The \textbf{Paraphrases} set consists of Combo instructions from the training set reformulated with novel vocabulary and syntax. The \textbf{New Objects} set introduces new combinations of Atomic goals that appeared during training but never occurred together in a single instruction. This structure allows FOCUS to assess both robustness to linguistic variation and compositional generalization.

\subsection{Experiments Setup}
\label{subsection:setup}

% TODO-new: hyperparamets to Appendix!
In our pipeline, we generate plans using Qwen2.5-14B-Instruct\footnote{\url{https://huggingface.co/Qwen/Qwen2.5-14B-Instruct}}, fine-tune it for one epoch with DPO ( \( \beta = 0.5 , \text{lr}= 1 \times 10^{-5} \) ) to stabilize local updates, and then train policies with PPO-T (\( \text{lr}=0.001,  \varepsilon =0.02 \)) and Skill Curriculum Learning for 2.5B steps. We validate by executing 10 plans across 50 seeds to assess robustness. Two full cycles were conducted, with evaluations before and after LLM fine-tuning, and results compared against baselines at 2.5B and 5B steps (Figure~\ref{fig:si_perfomance}). Additional hyperparameters are described in more detail in Appendix \ref{app:hyperparams}.

\begin{figure*}[!ht]
    \centering
    \includegraphics[width=1\linewidth]{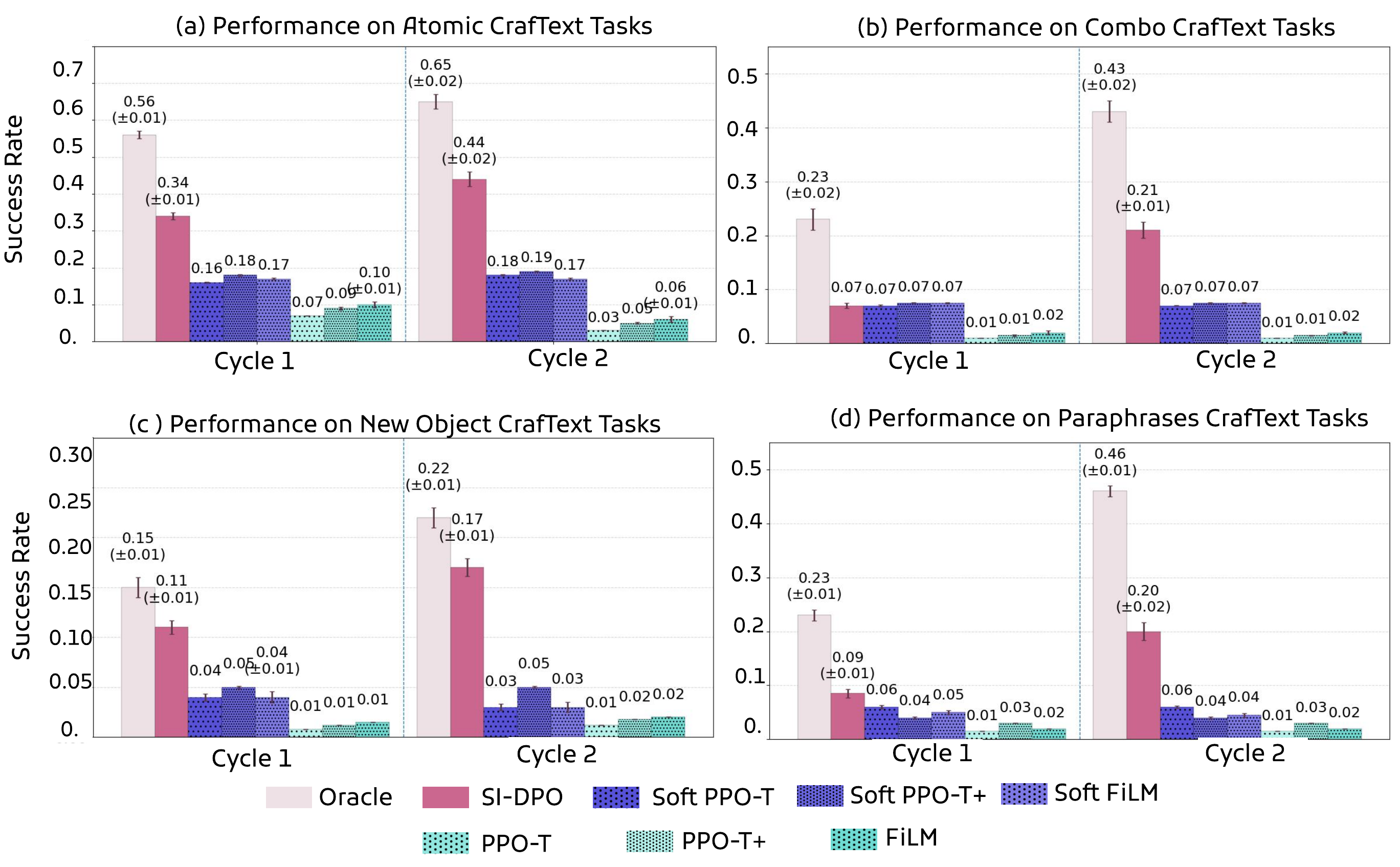}
    \caption{Comparison of SuperIgor and baseline performance on CrafText tasks (Atomic / Combo / New Objects / Paraphrases). All agents were evaluated at 2.5 billion steps (corresponding to the first cycle in the SuperIgor approach) and 5 billion steps (corresponding to the second cycle).}
    \label{fig:si_perfomance}
\vspace{-10px}

\end{figure*}
\subsection{Baselines}
\label{subsection:baselines}

For our comparative analysis, we use several established baselines from the original CrafText study~\citep{volovikova2025craftext}.
PPO-T (Text-Augmented PPO) augments PPO with textual grounding: instructions are encoded using a frozen DistilBERT [CLS] embedding, concatenated with CNN-based visual features, and processed by a GRU to maintain temporal context. PPO-T+ (Plan-Augmented PPO) extends this by first translating each instruction into a structured plan with GPT-4, and then providing the agent with a plan embedding instead of the raw instruction.

FiLM~\citep{perez2018film} offers an alternative integration of language and vision. Here, instruction embeddings generate parameters that modulate CNN outputs via Feature-wise Linear Modulation layers, allowing textual context to directly shape visual feature processing.

To ensure consistency, all baselines follow a strict protocol requiring the DONE action to signal task completion, with success only counted when both the instruction is satisfied and DONE invoked. We also evaluate an \textit{Auto-DONE} (\textit{Soft-}) variant, where episodes terminate automatically upon completion, and include an Oracle agent trained with PPO-T and Skill Curriculum Learning on human-written ground-truth plans.

\subsection{Experimental Results}

\textbf{RQ1. Effectiveness and Generalization of Auto-Generated Plans in the SuperIgor Pipeline} 

\textbf{a) Auto-generated plans train agents far more effectively than instruction-only baselines.} On Atomic tasks (Figure \ref{fig:si_perfomance}(a)), SuperIgor agent reach \(0.34\text{--}0.44\), compared to only \(0.10\text{--}0.19\) for instruction-only RL baselines. Oracle remains higher at \(0.56\text{--}0.65\), but the SuperIgor \(\rightarrow\) Oracle gap (\(\approx 0.20\)) is much smaller than the Baselines \(\rightarrow\) SuperIgor gap (\(\approx 0.25\text{--}0.30\)), clearly showing the value of plan supervision.  
On Combo tasks (Figure \ref{fig:si_perfomance}(b)), SuperIgor achieves \(0.21\), outperforming baselines at \(0.08\), while Oracle reaches \(0.43\). The wider gap to Oracle here can be explained by the fact that SI agents must simultaneously learn up to 20 alternative plans, whereas Oracle is trained on a single expert-aligned plan, which simplifies optimization. 

We further compare against standalone LLM-based agents (Appendix~\ref{app:SI_vs_LLM}) and observe that they fail to handle compositional instructions, with performance dropping sharply on Combo tasks. This highlights the importance of environment-grounded learning for effective long-horizon instruction following.

\textbf{b) Agents trained with auto-generated plans generalize on unseen goals better than those trained with Oracle plans.}  

On Combo tasks, Oracle achieves \(0.43\), while SuperIgor reaches \(0.21\). But on New Object tasks (Figure \ref{fig:si_perfomance}\text{(c)}), Oracle drops sharply to \(\approx 0.22\), while SI decreases more moderately to \(0.12\text{--}0.17\).  
Thus, although SI lags in absolute terms, its performance is more stable: the Oracle–SI gap shrinks from \(0.25\) on Combo to only \(0.05\text{--}0.10\) on New Object tasks. We attribute this stronger generalization precisely to the fact that SI agents learn from multiple alternative plans per instruction, which exposes them to richer variability during training.

\textbf{c) Agents trained with auto-generated plans do not lose performance when instructions are paraphrased.}  

Paraphrases reuse (Figure \ref{fig:si_perfomance} (d)) the same goals as in Combo tasks but are expressed in different linguistic forms. In Cycle~1, SI achieves \(0.07\) on Combo and \(0.09\) on Paraphrases. In Cycle~2, SI remains stable, with \(0.21\) on Combo and \(0.20\) on Paraphrases. This shows that SuperIgor agents can successfully transfer their learned strategies to differently worded instructions, maintaining performance even when the language of the goal changes.

\textbf{RQ2. Policy Training under Sparse Feedback}

\textbf{a) Skill Curriculum Learning significantly improves subtask acquisition under sparse feedback, compared to unstructured training}

%To answer this question, we evaluate the training process not just by final success rate, but by a more granular metric:

\begin{figure}[t]
    \centering
    \includegraphics[width=0.45\textwidth]{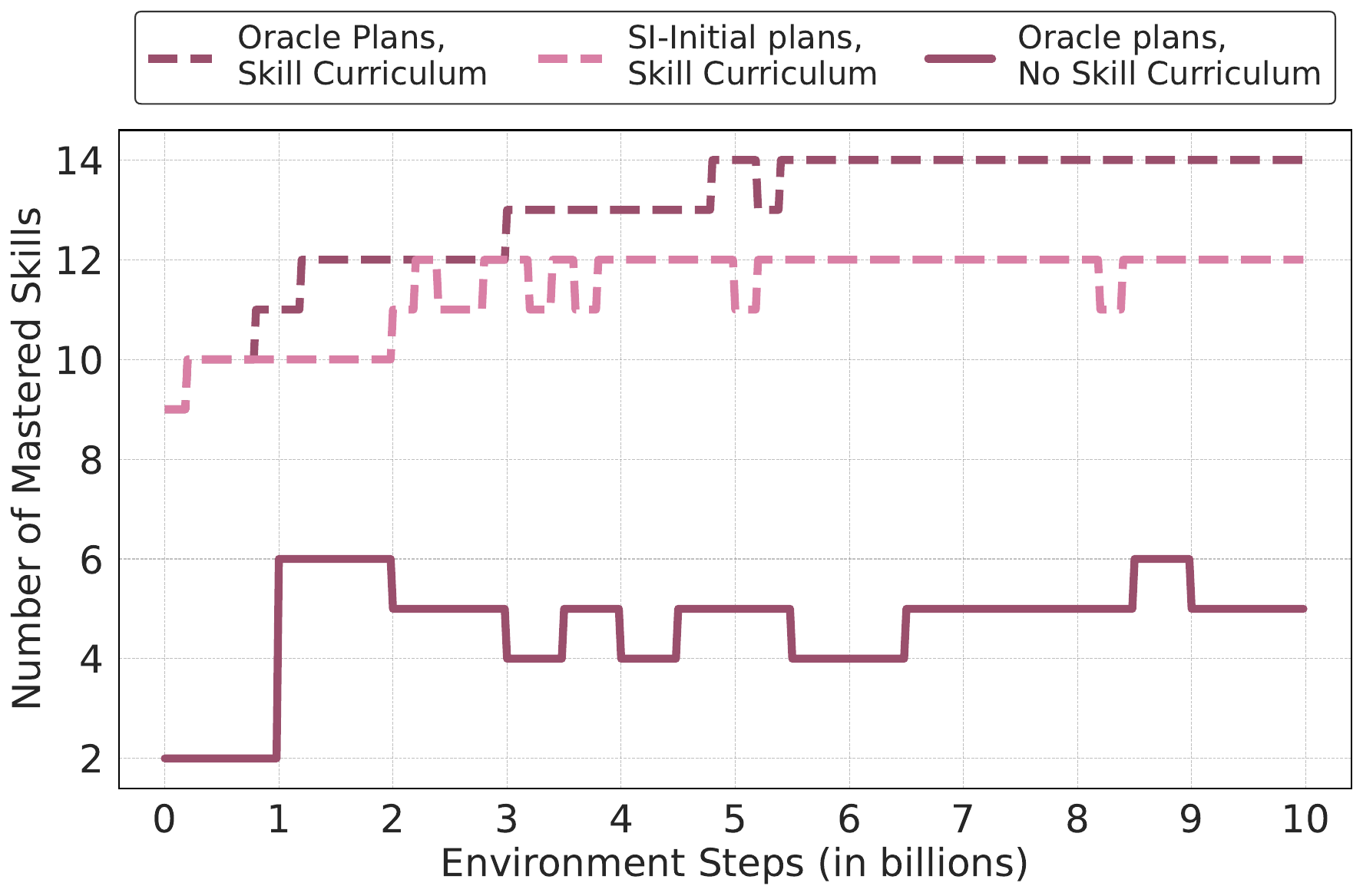}
    \caption{A comparative analysis of the number of mastered subtasks over 10 billion environment steps. }
    %The results highlight the critical role of the Skill Curriculum, as agents trained without it fail to learn, even with optimal Oracle Plans.
    \vspace{-10px}
    \label{fig:skill_mastery_comparison}
\end{figure}

 We evaluate the training process by the number of unique subtasks the agent masters over time. A subtask is considered "mastered" once its success rate surpasses a 70\% threshold. This metric provides a clearer insight into the agent's growing capabilities and its ability to handle compositional tasks. We compare three configurations, with the results visualized in Figure~\ref{fig:skill_mastery_comparison}.

The agent trained with \textbf{Skill Curriculum on Oracle Plans} sets a practical upper bound for performance. By the 10 billion step mark, it successfully masters \textbf{14 distinct subtasks}. It signifies that the agent has acquired almost the entire 'mining' technology tree: all the achievements from collecting wood to collecting iron. Furthermore, it demonstrates the ability to execute complex, combined instructions that require interleaving subtasks from different progression branches, such as eating, drinking, and collecting resources within a single, coherent plan.

Agent trained \textbf{on Oracle plans without the Skill Curriculum} perform worse with only mastered \textbf{5 basic subtasks}. Even with a flawless plan, the agent fails to learn without a structured progression that allows it to build foundational skills first. This finding confirms that Skill Curriculum helps to overcome sparse feedback problem and enables agent abilities to learn more subtasks.

\textbf{b) Plan quality is critical for effective learning under sparse feedback}

While curriculum learning is necessary for training under sparse rewards, its effectiveness strongly depends on the quality of the underlying plans. High-quality plans constrain the search space and provide a meaningful structure for skill acquisition, whereas poor plans can make learning intractable even with a well-designed curriculum.

Skill Curriculum with SI-Initial plans closely follows the Oracle trajectory, mastering \textbf{12 subtasks} within the same timeframe (compared to \textbf{14} with Oracle plans). This shows that plans generated by Qwen-14B-Instruct are a strong approximation to expert-designed ones and are sufficient to unlock most of the agent’s learning potential.

However, we observe that this performance is highly sensitive to the underlying language model used for plan generation; a detailed analysis is provided in Appendix~\ref{app:ontology_quality} and Appendix~\ref{app: ablation_llm}. When the initial plans are of lower quality, the curriculum alone is insufficient to overcome the sparse reward setting, and the agent fails to learn effectively.

% In conclusion, the curriculum is not just beneficial, it is \textit{critical} for meaningful skill acquisition in this environment. The ablation clearly shows that our Skill Curriculum Learning framework is the key enabler of learning, while our \textsc{SI-Initial} procedure generates plans of sufficient quality to unlock a significant portion of the agent's potential and a good baseline for furthermore plan generation improvement using SuperIgor framework. 

\textbf{RQ3. Agent Effectiveness with Iterative SuperIgor Cycles}

\textbf{a) Plan-following quality improves across cycles.}  
On Atomic tasks (training, Figure \ref{fig:si_perfomance}, (a)), SI-DPO increases from \(0.34\) in Cycle~1 to \(0.43\) in Cycle~2. On Combo tasks (training, Figure \ref{fig:si_perfomance}, (b)), SI-DPO grows from \(0.06\) in Cycle~1 to \(\approx 0.21\) in Cycle~2. On New Object tasks (testing, Figure \ref{fig:si_perfomance}, (c)), SI-DPO declines only slightly from \(\approx 0.21\) to \(0.12\text{--}0.17\), showing that performance improves with additional SuperIgor cycles on both training and testing setups and remains relatively stable when moving to unseen goals.

\textbf{b) Plan reprioritization under DPO illustrates the process by which language models are incrementally grounded in the agent’s behavior and the underlying environment mechanics.}.  
The re-ranking visualization (Appendix \ref{app:reprioritization}, Figure \ref{fig:dpo_reparam}) shows how plans shift across SFT, DPO-C1, and DPO-C2. Success Rates range from \(0.68\) to \(0.86\). A plan with SR \(=0.86\) steadily climbs to the top across cycles, while weaker plans with SR \(\approx 0.68\) remain consistently at the bottom. These changes are gradual rather than abrupt, suggesting that DPO provides a soft grounding signal that progressively aligns plan priorities with the agent’s execution success. This interpretation is further supported by the ablation study (Section \ref{sec:ablation}), which shows that integrating DPO accelerates agent learning in the second cycle, indicating that gradual plan re-prioritization translates into more effective behavioral grounding. We also compare DPO with GRPO (Appendix \ref{app: dpo_with_grpo}) and observe that reward-based optimization is less effective in this setting, likely due to its sensitivity to reward scale variability across instructions, making it less aligned with the plan ranking objective.

\section{Ablation Study}
\label{sec:ablation}

\begin{table*}[!ht]
\small
\vspace{-6px}
 \caption{Ablation study of the SuperIgor framework, measuring agent SuccessRate on the Atom subset of the CrafText dataset across two training cycles}
 \label{tab:ablation}
 
\centering
\renewcommand{\arraystretch}{1.}
\renewcommand{\tabularxcolumn}[1]{>{\centering\arraybackslash}m{#1}}
\begin{tabularx}{\textwidth}{XXXX|XX}
\toprule
\textbf{Ontology} & \textbf{Curriculum} & \textbf{DPO} & \textbf{SFT} &
\textbf{Cycle-1} & \textbf{Cycle-2} \\
\midrule
\ding{55} & \checkmark & \checkmark & \checkmark & 0.06 & N/A \\
\checkmark & \ding{55} & \checkmark & \checkmark & 0.08 & N/A \\
\checkmark & \checkmark & \ding{55} & \checkmark & 0.34 & 0.39 \\
\checkmark & \checkmark & \checkmark & \ding{55} & 0.25 & 0.13 \\
\checkmark & \checkmark & \checkmark & \checkmark & \textbf{0.35} & \textbf{0.45} \\
\bottomrule
\end{tabularx}
\vspace{-10px}
\end{table*}

\subsection{Component Interaction Ablation}
 \label{app: aplation}
To quantify the contribution of each module of the SuperIgor framework, we conduct an ablation study in which individual components are removed from the training pipeline. We evaluate the influence of four factors: 
(\(1\)) Ontology-Based Training Plan Generation, 
(\(2\)) Curriculum design in the RL stage, 
(\(3\)) LLM plan-model pretraining (SFT), and 
(\(4\)) DPO finetuning based on RL agent performance signals.
Table \ref{tab:ablation} presents the results of this experiment, where we measure the SuperIgor agent’s SuccessRate on the Atom subset of the CrafText instruction dataset. The analysis of the results yields two central findings.

\textbf{(1) Curriculum is effective only when paired with high-quality, ontology-structured plans.}
Although full-cycle results may suggest that gains are driven by curriculum learning, the ablation shows that curriculum is effective only when combined with ontology-guided plan generation. Without ontology—i.e., without structured, hierarchical plans—the curriculum lacks a meaningful ordering signal and fails to improve performance (Cycle-\(1\): \(0.06\)). In contrast, ontology-based plans encode a natural hierarchy of goals, enabling a principled progression from simple to complex tasks. This alignment between plan structure and staged learning makes the curriculum operative, increasing Cycle-\(1\) performance from \(0.06\) to \(0.35\).
% Although a full-cycle evaluation may give the impression that the primary gains come from curriculum learning, the results of this ablation study show that its effectiveness emerges only in combination with ontology-guided plan generation. Without ontology (i.e., without structured, hierarchical plans), the curriculum has no meaningful ordering signal and fails to provide improvement: Cycle-\(1\) performance drops to \(0.06\) when ontology is removed.

% Ontology-based plans, however, naturally encode a hierarchy of instructions and goals, enabling a principled progression from simpler to more complex targets. This hierarchical structure is precisely what makes a curriculum implementable: the RL agent can first master low-complexity goals and then gradually advance to more difficult ones. When ontology is present, this alignment between plan structure and staged learning produces large gains, improving Cycle-\(1\) performance from \(0.06\) (no curriculum) to \(0.35\) (with curriculum).

\textbf{(2) DPO improves the RL agent by learning to prioritize plans that lead to higher-quality behavior.}
Unlike SFT, which is trained to reproduce the ontology-induced distribution of plans, DPO directly leverages RL performance as a preference signal: it learns to rank plans higher when they empirically yield better agent behavior. Removing DPO results in weaker prioritization: the RL agent reaches only \(0.39\) in Cycle-\(2\) without DPO, compared to \(0.45\) when DPO is included. Thus, DPO systematically shifts the plan distribution toward behaviorally effective plans, accelerating and amplifying the RL agent's improvement across cycles.

\subsection{Compute-Matched Ablations}

% To isolate planner adaptation from pure policy optimization, we run compute-matched ablations (Table \ref{tab:compute}) extending PPO to 15B steps with a fixed planner (no DPO). Success Rate improves early (0.26$\rightarrow$0.32 at 5B) but then saturates around 0.30--0.31 despite 3$\times$ more compute, indicating that scaling the policy alone cannot overcome a fixed plan distribution under sparse rewards. At 5B steps, we compare: (i) longer training without DPO, (ii) no DPO with reward-based plan filtering, and (iii) the full SuperIgor pipeline with DPO. Filtering boosts training performance (0.30$\rightarrow$0.34) but harms generalization (0.11$\rightarrow$0.10), suggesting overfitting, while SuperIgor improves both training (0.36) and generalization (0.17). Overall, gains arise from planner--policy co-adaptation rather than increased RL compute.

To isolate the effect of planner adaptation from pure policy optimization, we conduct compute-matched ablations where PPO training is extended up to 15B environment steps while keeping the planner fixed (no DPO). We observe that the Success Rate improves during early training (0.26 $\rightarrow$ 0.32 at 5B steps) but saturates thereafter, stabilizing around 0.30--0.31 despite a 3$\times$ increase in compute. This indicates that scaling the policy alone is insufficient to overcome the limitations imposed by a fixed plan distribution, consistent with the sparse-reward challenges discussed in Section~3.

We further compare three compute-matched variants at 5B steps: (i) no DPO with longer training, (ii) no DPO with reward-based plan filtering, and (iii) the full SuperIgor pipeline with DPO. While filtering improves training performance (0.30 $\rightarrow$ 0.34), it degrades generalization on New Objects (0.11 $\rightarrow$ 0.10), suggesting overfitting to a narrower plan subset. In contrast, SuperIgor (SI-DPO) improves both training (0.36) and generalization (0.17), demonstrating that iterative planner updates align the plan distribution with the policy's learning dynamics. These results confirm that performance gains arise from planner--policy co-adaptation rather than increased RL compute alone.

\begin{table}[!ht]
\small
\centering
\caption{Compute-matched comparison of static and adaptive planner variants.}
\setlength{\tabcolsep}{3.5pt}
\begin{tabular}{lcc}
\toprule
Model & Train Combo & Test New Objects \\
\midrule
No DPO, longer training & 0.30 & 0.11 \\
No DPO, reward filtering & 0.34 & 0.10 \\
SuperIgor (SI-DPO) & \textbf{0.36} & \textbf{0.17} \\
\bottomrule
\end{tabular}
\label{tab:compute}
%\vspace{-10px}
\end{table}

\section{Conclusion}

In this work, we introduced SuperIgor, a framework for training agents to follow complex natural-language instructions in dynamic multimodal environments. It combines language-model-based planning with reinforcement learning, enabling effective instruction-following without predefined subtasks or explicit subgoal verification. Ontology-based plan generation provides structure, supporting both Skill Curriculum Learning for sparse rewards and cyclic DPO-based adaptation to prioritize effective plans. Together, these components allow SuperIgor to outperform instruction-only baselines, achieve near-Oracle generalization on out-of-distribution tasks, and remain robust to instruction paraphrasing.

% In this work, we introduced SuperIgor, a framework for training agents to follow complex natural-language instructions in dynamic, multimodal environments. We showed that effective instruction-following can be learned by integrating language-model-based planning with reinforcement learning, even in the absence of predefined subtasks and without explicit verification of intermediate subgoal completion.This is achieved through the integration of ontology-based plan generation, which provides the structural foundation for learning and enables both Skill Curriculum Learning to handle sparse and delayed rewards and cyclic DPO-based language model adaptation that prioritizes effective plans and accelerates agent training. Together, these components enable SuperIgor to outperform instruction-only baselines, generalize near Oracle on out-of-distribution tasks, and remain robust to instruction paraphrasing.

\section*{Limitations}
While SuperIgor demonstrates a scalable approach to integrating language-model-based planning with reinforcement learning, it has several limitations. First, the method depends on the quality of the initial plan structure induced by the language model. The ontology graph is generated fully automatically, without human supervision, which may lead to redundant, inconsistent, or cyclic dependencies between subtasks. In practice, we mitigate this issue empirically by evaluating different language models and selecting those that produce stable and coherent plan graphs (Appendix \ref{app: ablation_llm}); our experiments show that several mid-sized models already perform well in this setting.Nevertheless, robustness to poorly structured plan spaces remains a limitation of the current approach. 

In addition, the interaction between plan quality and policy learning is not always interpretable. While our validation protocol provides a clear relative preference signal between candidate plans for the same instruction (via repeated evaluation across seeds), it can still be difficult to diagnose training failures at the system level. In particular, when learning stalls, it is unclear whether the bottleneck is caused by limitations of the current policy optimization and exploration dynamics or by systematic issues in the generated plan space (e.g., missing prerequisites or overly difficult decompositions).

\bibliography{si}

 \newpage
\clearpage
\appendix

%\section*{Appendix}

\section{Plans Generation:candidates generation}
\label{app:zoe}

\begin{figure*}[t]
    \centering
    \includegraphics[width=0.9\linewidth]{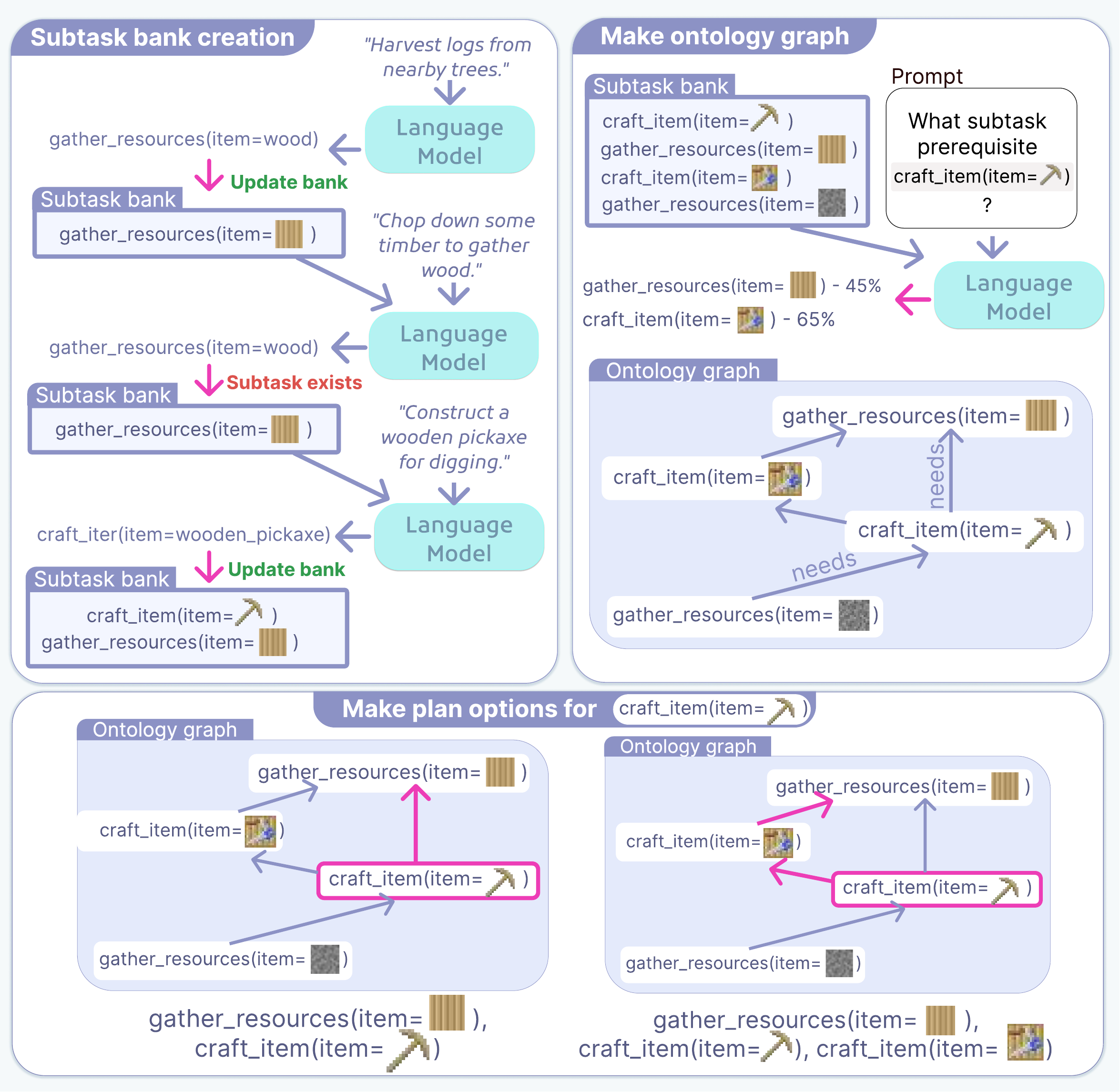}
    \caption{Training plans generation}
    \label{fig:plans_generation}
\end{figure*}

In our plan extraction method the goal is to elicit the model’s hypotheses about the dependencies between subtasks in the environment. In the first step we construct a subtask bank $B$, i.e., the set of all candidate subtasks derived from the instruction set. For each instruction $I \in \mathcal{D}$, we prompt the language model $f_{\text{LLM}}$ to generate a goals plan $\mathcal{P}[I]$, i.e., the set of goal subtasks directly required by the instruction. The model is provided with the current contents of the subtask bank $B$, which encourages reuse of already known subtasks and reduces the introduction of redundant synonyms. If the generated goals contain subtasks not yet present in $B$, they are added. At the initial iteration the bank is empty, so all subtasks generated by the model are included. The complete process is summarized in Algorithm~\ref{alg:zoe-goals}.

Once a sufficiently rich subtask bank $B$ has been established, ontological dependencies between subtasks are extracted. For each target subtask $t \in B$, the language model is queried multiple times to determine which elements from $B$ are required for the completion of $t$. For every candidate dependency $(r \to t)$, its probability is estimated as
\[
P(r \to t) = \frac{k_t}{N},
\]
where $k_t$ denotes the number of times subtask $r$ was identified as necessary for $t$ and $N$ is the number of queries. To filter out spurious associations, the Wilson confidence interval is applied to the resulting probabilities. The procedure is carried out in two passes: first over the entire bank $B$, and then restricted to the subtasks previously identified as relevant, which refines the weighting of relations. The final output is an ontology graph $G=(V,E)$ that encodes the model’s hypothesized structure of interrelations among subtasks. The full procedure is summarized in Algorithm~\ref{alg:zoe-ontology}.

After constructing the ontology $G=(V,E)$, each goal plan $\mathcal{P}[I]$ is expanded with its dependencies. For every subtask $s \in \mathcal{P}[I]$, we recursively collect all prerequisites in $G$. The union of these subtasks with the original goals defines the plan’s vertices, which are then topologically sorted so that prerequisites precede dependents. The result is a linearized plan $P$ containing the goals and all supporting subtasks (Algorithm~\ref{alg:zoe-final-plan}).

% \vspace{-10px}
\begin{algorithm}[H]
\footnotesize
\caption{Subtask Bank Update}
\label{alg:zoe-goals}
\begin{algorithmic}[1]
\REQUIRE Instruction stream $\mathcal{D}$, language model $f_{\text{LLM}}$
\ENSURE Subtask bank $B$, goals plans $\mathcal{P}$

\STATE Initialize subtask bank $B \gets \emptyset$
\STATE Initialize goals plans $\mathcal{P} \gets \emptyset$

\FOR{each instruction $I \in \mathcal{D}$}
    \STATE Identify goal subtasks conditioned on $B$:
    \[
        S \gets f_{\text{LLM}}(I, B)
    \]
    \FOR{each subtask $s \in S$}
        \IF{$s \notin B$}
            \STATE $B \gets B \cup \{s\}$
        \ENDIF
    \ENDFOR
    \STATE Goals plan for $I$: $\mathcal{P}[I] \gets S$
\ENDFOR

\STATE \textbf{return} $B, \mathcal{P}$
\end{algorithmic}
\end{algorithm}
% \vspace{-15px}

\vspace{-10px}
\begin{algorithm}[H]
\footnotesize
\caption{Ontology Construction}
\label{alg:zoe-ontology}
\begin{algorithmic}[1]
\REQUIRE Subtask bank $B$, language model $f_{\text{LLM}}$, queries per pass $N$, threshold $\tau$
\ENSURE Ontology graph $G=(V,E)$

\STATE Initialize counts $count(r,t) \gets 0$ for all $r,t \in B, r \neq t$

\FOR{each target subtask $t \in B$}
    \FOR{two passes}
        \STATE Define candidate set $C$: 
        \[
            C \gets 
            \begin{cases}
                B \setminus \{t\}, & \text{pass 1} \\
                \{r \in B : count(r,t) > 0\}, & \text{pass 2}
            \end{cases}
        \]
        \FOR{$i = 1 \dots N$}
            \STATE Query prerequisites:
            \[
                R \gets f_{\text{LLM}}(t, C)
            \]
            \FOR{each $r \in R$}
                \STATE $count(r,t) \gets count(r,t) + 1$
            \ENDFOR
        \ENDFOR
    \ENDFOR
\ENDFOR

\STATE Initialize edge set $E \gets \emptyset$
\FOR{each pair $(r,t)$}
    \STATE Compute probability:
    \[
        \hat{p}(r \to t) = \frac{count(r,t)}{N}
    \]
    \STATE Compute Wilson lower bound $LB(\hat{p}, N)$
    \IF{$LB \geq \tau$}
        \STATE $E \gets E \cup \{(r \to t)\}$
    \ENDIF
\ENDFOR

\STATE \textbf{return} $G = (V=B, E)$
\end{algorithmic}
\end{algorithm}

\newcommand{\Call}[2]{\textsc{#1}(#2)} \begin{algorithm}[H] 
\caption{ Final Plan Generation from Ontology} \label{alg:zoe-final-plan} 
\begin{algorithmic}[1] 
\REQUIRE Instruction $I$, goals mapping $\mathcal{G}$, goals plan $\mathcal{P}$, ontology $G=(V,E)$ 
\ENSURE Final plan $P$ 
\STATE Retrieve goal subtasks: $S \gets \mathcal{G}[I]$ 
\STATE Initialize plan vertex set: $U \gets S$ \FOR{each $s \in S$} 
\STATE Expand prerequisites via ontology: \[ D \gets \Call{PrereqClosure}{s, G} \] \STATE $U \gets U \cup D$ 
\ENDFOR 
\STATE Extract induced subgraph: $G_U \gets G[U]$ 
\STATE Topologically sort $G_U$ to obtain ordered plan $P$ 
\STATE \textbf{return} $P$ 
\end{algorithmic} 
\end{algorithm}

\section {Additional Experiments}

\subsection{SuperIgor vs. SOTA LLMs}
\label{app:SI_vs_LLM}

\begin{table}[h]
\centering
\caption{Model performance comparison}
\begin{tabular}{lcc}
\hline
Model & Atom & Combo \\
\hline
SI\_pixel (Cycle 1) & 0.56 & 0.23 \\
Claude Sonnet 4.6 & 0.24 & 0.05 \\
OpenAI GPT-4.1 & 0.14 & 0.04 \\
Qwen/QWQ-32B & 0.22 & 0.03 \\
Gemma-27B & 0.06 & 0.02 \\
\hline
\label{tab:models_comparison}
\end{tabular}
\end{table}

We additionally evaluate standalone LLM-based agents under the same \textit{Atom} and \textit{Combo} settings (Table~\ref{tab:models_comparison}). While several models achieve moderate performance on atomic tasks, all of them exhibit a sharp degradation on compositional instructions, with success rates dropping close to zero in the \textit{Combo} setting. This trend holds even for reasoning-oriented models such as QWQ-32B, indicating that improvements in single-step reasoning do not directly translate to effective long-horizon planning.

In contrast, the RL-trained SuperIgor agent maintains substantially higher performance and degrades more gracefully under composition. These results suggest that purely LLM-based approaches struggle with compositional planning and long-horizon credit assignment, consistent with observations in CrafText and related embodied benchmarks . We also observe sensitivity of LLM agents to prompt design and observation encoding, further limiting their stability and reproducibility. Overall, this comparison highlights the necessity of environment-grounded learning for reliable instruction following in complex settings.

\subsection{Ablation Study: Skill Mastery Threshold}
\label{app:ablation_sr_threshold}

\begin{figure}[t]
    \centering
    \includegraphics[width=0.9\linewidth, height=0.2\textheight]{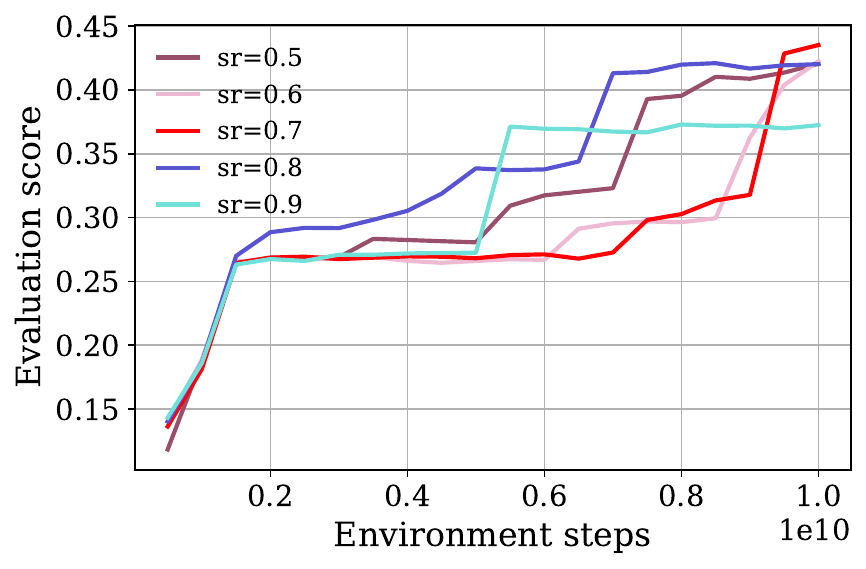}
    \caption{Ablation of the skill-mastery threshold $\tau$. The plot shows evaluation scores on the Atomic and Combo tasks during training for different $\tau$ values.}
    \label{fig:ablation_sr_threshold}
\end{figure}
We conducted an ablation study to analyze the sensitivity of the Skill Curriculum Learning to the mastery threshold parameter $\tau$. Figure~\ref{fig:ablation_sr_threshold} presents the final performance of the Skill Curriculum Learning agent after 10 billion environment steps in CrafText-Symbolic configuration for different values of $\tau$ ranging from $0.5$ to $0.9$.

The results demonstrate that $\tau=0.7$ provides an optimal balance for curriculum progression. We hypothesize that lower thresholds ($\tau=0.5$) allow the agent to progress too quickly to complex skills before achieving reliable proficiency, while higher thresholds ($\tau=0.9$) cause the agent to spend excessive time perfecting basic skills, slowing overall learning. The $\tau=0.7$ value strikes an optimal balance between progression speed and skill reliability.

\subsection{Analysis of Ontology Quality}
\label{app:ontology_quality}

\begin{figure}
    \centering
    \includegraphics[width=1\linewidth]{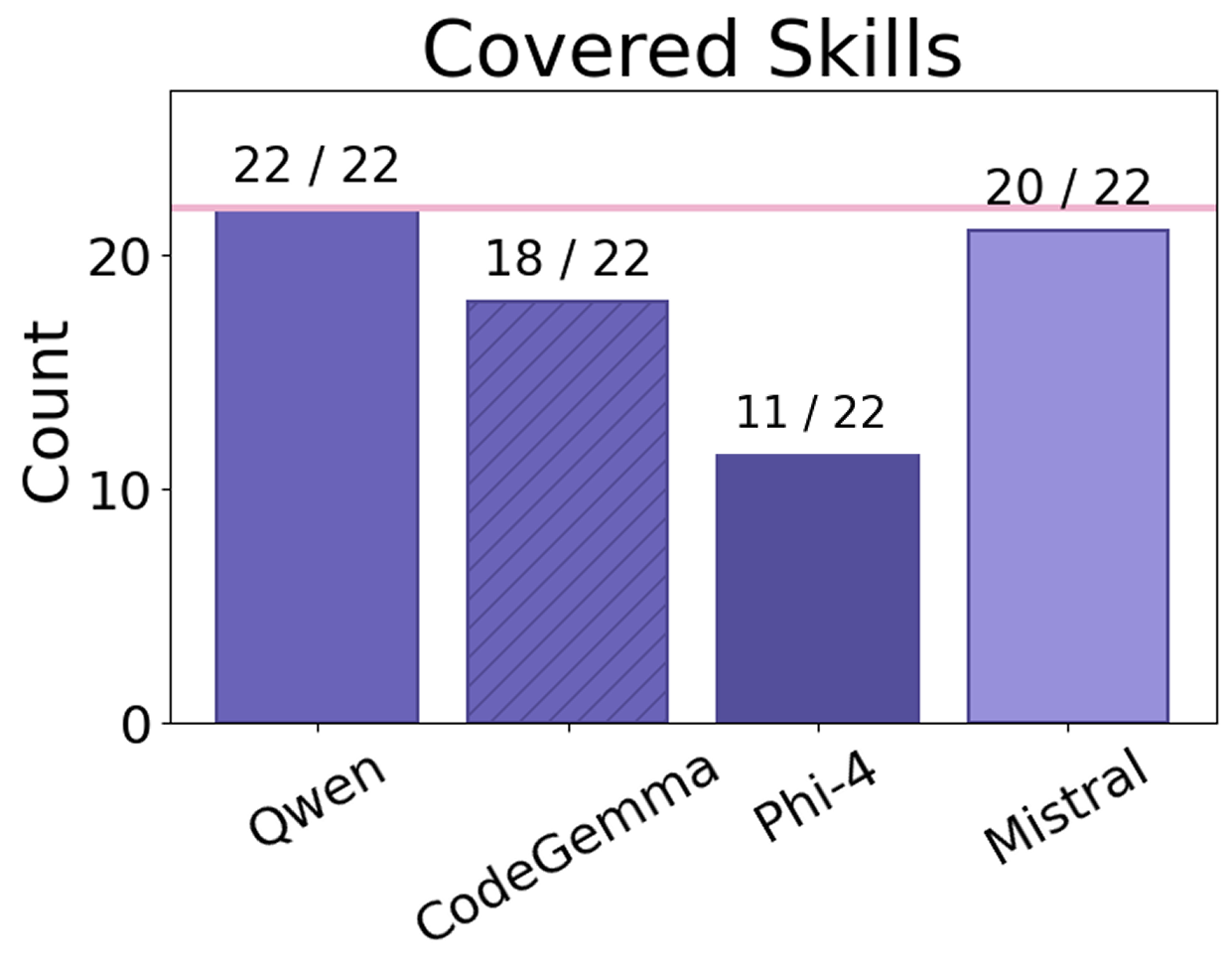}
    \caption{Coverage of reference subtasks (out of 23) across different LLMs. Higher is better.}
    \label{fig:covered_skills}
\end{figure}

We investigate how the quality of subtask extraction affects downstream performance. In particular, we focus on three factors: (i) oversampling of redundant or synonymous subtasks (\textit{Oversampling}), (ii) missing core subtasks (\textit{Missing Skills}), and (iii) the presence of spurious (non-classified) subtasks (\textit{Spurious Skills}). To evaluate this, we construct a reference set by sampling $\sim$50 validation instructions and manually annotating their core subtasks, resulting in 23 unique skills. Generated subtasks are matched against this set using cosine similarity, and we compute the corresponding metrics. Coverage results are shown in Figure~\ref{fig:covered_skills}, while the remaining metrics are summarized in Table~\ref{tab:ontology_quality}.

We observe a clear relationship between ontology quality and downstream performance. As shown in Figure~\ref{fig:covered_skills}, models with higher coverage recover a larger fraction of the reference subtask set. At the same time, Table~\ref{tab:ontology_quality} shows that low-performing models tend to exhibit higher oversampling and a larger number of missing and spurious subtasks. 

In particular, oversampling leads to an inflated and ambiguous subtask space, while missing skills result in incomplete plans that cannot be reliably executed. In contrast, structural issues such as cyclic dependencies, while undesirable, can be mitigated algorithmically (e.g., via cycle detection and pruning) and appear less critical in practice.

Overall, these results indicate that downstream performance is primarily driven by the precision and completeness of subtask extraction, highlighting the importance of controlling subtask granularity and redundancy for stable and effective training.

\begin{table}[h]
\centering
\caption{Ontology quality metrics across LLMs (lower is better): Over.~-- oversampling, Missing~-- missing skills, Spurious~-- non-classified skills.}
\label{tab:ontology_quality}
\small
\begin{tabular}{lccc}
\hline
Model & Over. $\downarrow$ & Missing $\downarrow$ & Spurious $\downarrow$ \\
\hline
Qwen & 1 & 2 & 1 \\
CodeGemma & 1 & 5 & 3 \\
Phi-4 & 25 & 8 & 6 \\
Mistral & 59 & 2 & 9 \\
\hline
\end{tabular}
\end{table}

\subsection{Ablation Study: Choice of LLM for Ontology and Training-Plan Generation}
\label{app: ablation_llm}

To understand how the choice of language model affects the quality of the ontology and the generated training plans we conducted an ablation study comparing several families of LLMs. For each model, we regenerated both the ontology and the full training dataset (plans), and then trained an RL agent using our Skill Curriculum Learning procedure.

Table~\ref{tab:llm_ablation} reports the agent’s success rate on the training split under different planner models. The experiment includes models from the Qwen and Gemma families, as well as the larger \texttt{microsoft/NextCoder-32B} model.

\begin{table*}[h]
\small
% \vspace{-4px}
\caption{Ablation on the choice of LLM used for generating both ontology and training plans. We report success rate on the training set.}
\label{tab:llm_ablation}
\centering
\renewcommand{\arraystretch}{1.2}
\begin{tabular}{l|cccccccc}
\toprule
% \textbf{LLM} & Soft-PPO-T+ & Soft-PPO-T & Soft-Film & Qwen1.5-32B & NextCoder-32B & Qwen1.5-14B & Gemma-12B & Qwen-7B \\
\textbf{LLM} & Qwen1.5-32B & NextCoder-32B & Qwen1.5-14B & Gemma-12B & Qwen-7B \\
\midrule
% \textbf{SR (Train)} & 0.13 & 0.12 & 0.12 & 0.00 & 0.26 & 0.35 & 0.14 & 0.22 \\
\textbf{SR (Train)} & 0.43 & 0.26 & 0.35 & 0.14 & 0.22 \\
\bottomrule
\end{tabular}
\end{table*}

% \vspace{-10px}
\begin{figure}[t]
    \centering
    \includegraphics[width=\linewidth]{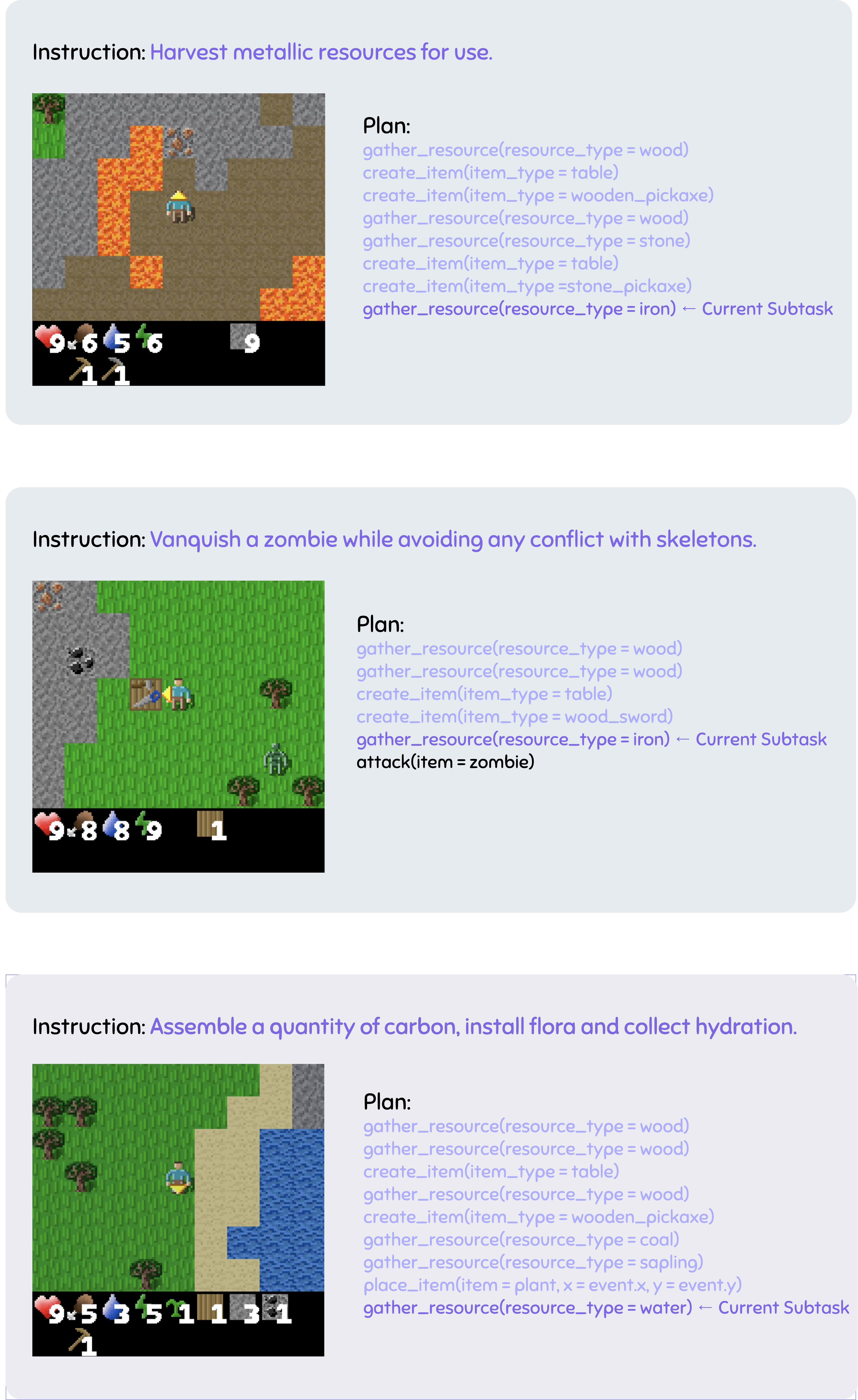}
    \caption{Example of instructions and corresponding plans}
    \label{fig:instruction_examples}
\end{figure}
% \vspace{-10px}

% \begin{wrapfigure}{r}{0.48\textwidth}
% \vspace{-10px}
%     \centering
%     \vspace{-5px}
%     \includegraphics[width=1.\linewidth]{figures/instruction_examples.pdf}
%     \caption{Example of instructions and corresponding plans}
%     \label{fig:instruction_examples}
% \end{wrapfigure}

\textbf{(1) Larger models do not necessarily produce better ontologies or plans.}
Although one may expect the largest models to generate the most structured plans, but NextCoder-32B performance is surpassed by significantly smaller Qwen models. Qwen-32B yields the highest performance (0.43), and even Qwen-7B outperforms Gemma-12B, indicating that model family and training specialization matter more than raw parameter count.

\textbf{(2) Qwen models produce more stable and semantically consistent plan structures.}
Models from the Qwen family demonstrate higher robustness in generating hierarchical task decompositions that align with our ontology constraints. This leads to more reliable curriculum construction and more effective RL training.

\textbf{(3) Some widely used LLMs fail to benefit from the alignment stage.}
We also conducted experiments with several other well-known models, including \textit{microsoft/phi-4}, \textit{mistralai/Mistral-7B-Instruct-v0.2}, and \textit{openai/gpt-oss-20b}, and found that the alignment stage does not provide any measurable benefit for them. Despite explicit prompt constraints on which subtasks should be used, these models tend to generate large numbers of synonymously similar subtasks. Consequently, the set of goals that the agent must recover becomes even larger than when instructions are provided directly, rendering it impractical to run the full pipeline with these models.

\subsection{DPO vs.\ GRPO for Planner Fine-Tuning}
\label{app: dpo_with_grpo}

We additionally compare DPO and GRPO as alternative methods for planner fine-tuning. While GRPO and PPO are in principle applicable to this setting, recent work suggests that GRPO can provide a stronger and more stable optimization signal for reward-based language model training. To evaluate whether this also holds in our setup, we replace DPO with GRPO in the planner fine-tuning stage and measure downstream policy performance on the validation split. Results are reported in Table~\ref{tab:dpo_vs_grpo}.

We find that DPO achieves better performance than GRPO in our setting, improving success rate from 0.18 to 0.21 while also reducing the number of unsolved instructions from 175 to 166. We attribute this to the structure of the planner objective: although dense execution scores are available, the planner is not required to optimize absolute reward values directly, but rather to rank alternative plans for the same instruction. Since reward magnitudes may vary substantially across instructions, direct reward optimization is less well aligned with this goal. In contrast, DPO explicitly optimizes pairwise preferences, which better matches the plan-selection problem considered in our framework.

\begin{table}[h]
\centering
\caption{Comparison of planner fine-tuning methods on the validation split. Higher SR is better; lower unsolved count is better.}
\label{tab:dpo_vs_grpo}
\small
\begin{tabular}{lcc}
\hline
Model & SR $\uparrow$ & Unsolved $\downarrow$ \\
\hline
Original (pretrained) & 0.11 & 234 \\
DPO & 0.21 & 166 \\
GRPO & 0.18 & 175 \\
\hline
\end{tabular}
\end{table}

\section{CrafText}
%\vspace{30px}
\label{app:craftext_description}

To provide a concrete example of our method, Figure \ref{fig:instruction_examples} visualizes the agent's state at a single timestep. The CrafText environment, shown on the left, is a dynamic grid-world where the agent must gather resources, craft items, and navigate diverse terrains to survive and complete tasks.

The core of our approach lies in the hierarchical decomposition of complex goals. As shown on the right, a high-level instruction, which may be ambiguous or require long-term planning (e.g., "Craft an iron pickaxe."), is first translated into a deterministic, multi-step plan. Each step in this plan constitutes a distinct subtask.

Crucially, the agent's policy is not conditioned on the entire plan. Instead, it focuses solely on the currently active subtask. This transforms a challenging long-horizon problem into a more tractable sequence of short-horizon tasks. The agent's objective at any moment is to complete the highlighted subtask and then invoke the DONE action. For example, optimal agent can choose DONE action based on the inventory state (when completing subtasks such as collecting resources and crafting items), player status (for subtasks that are related to eating, drinking or sleeping) or map state (for subtasks such as placing blocks).

Upon successful completion, the framework provides the next subtask in the sequence, guiding the agent through the overall plan until the final goal is achieved.

% For our work we used a variation of Easy Craftext dataset EASY-STRICT, which introduces more strict instruction completition protocol. The structure of the dataset is as follows:

We utilize the \textbf{FOCUS} instruction dataset described in the main text. The structure of the dataset is as follows:
    
Training Set:
    \begin{itemize}
        \item  Atomic: Single, indivisible goals (e.g., "Craft a furnace").
        \item  Combo: Sequences of multiple atomic goals (e.g., "Craft a furnace and then collect wood").
        \item  Crucially, each instruction in the training set also has a paraphrased version to encourage linguistic robustness from the start.
    \end{itemize}

Test Sets (Out-of-Distribution):
    \begin{itemize}
        \item  Paraphrases: Contains the same underlying goals as the Combo training set, but expressed with novel vocabulary and syntax. This tests robustness to linguistic variation.
        \begin{itemize}
            \item  Training Combo: "Consume beef and create a stone pickaxe."
            \item Test Paraphrase: "Eat steak and forge a stone pickaxe." or "Devour cow meat and create a stone pickaxe."
        \end{itemize}
        \item  New Objects: Introduces new combinations of atomic goals that appeared during training but never occurred together in a single instruction in the training set. This directly tests compositional generalization. These instructions also come with their own paraphrases.
        \item  Training contained: "Consume beef" and "Forge a stone pickaxe" and "Forge a stone blade" as separate atomic or part of other combos.
        \item  Test New Object: "Consume beef and forge a stone blade." or "Eat cow meat and create a sword from stone."
    \end{itemize}

This structure allows us to rigorously dissect the agent's capabilities: learning from language (Atomic/Combo), generalizing to new phrasing (Paraphrases), and generalizing to new goal combinations (New Objects).

\section{Complete SuperIgor Training Pipeline}

The SuperIgor framework integrates multiple components that exchange specific inputs and outputs during training. Below we describe the key data flows between components:

\textbf{Component Interfaces:}

\begin{itemize} % [noitemsep]
    \item \textbf{LLM Planner ($f_{\text{LLM}}$)}
    \begin{itemize} % [noitemsep]
        \item \textbf{Input:} Instruction $I$
        \item \textbf{Output:} Candidate plan $\mathcal{P}$ consisting of a sequence of subtasks from subtask bank $\mathcal{B}$
    \end{itemize}
    \item \textbf{RL Policy ($\pi_{\theta}$)}
    \begin{itemize} % [noitemsep]
        \item \textbf{Input:} Environment observations $o_t$, current plan step DistilBERT [CLS] embedding $p_{\phi(t)}$ of a plan $\mathcal{P}$
        \item \textbf{Output:} Action $a_t$ from extended action space containing default Craftext actions and additional DONE action that gives the agent the next plan step embedding $p_{\phi(t + 1)}$ of a plan $\mathcal{P}$
    \end{itemize}
\end{itemize}

The complete training procedure integrating all components is summarized in Algorithm~\ref{alg:superigor_complete}

\begin{algorithm}[h]
\footnotesize
\caption{Complete SuperIgor Training Pipeline}
\label{alg:superigor_complete}
\begin{algorithmic}[1]
\REQUIRE 
\STATE Environment $\mathcal{E}$
\STATE Instruction dataset $\mathcal{D}_{\text{train}} = \{I_1, I_2, \dots, I_N\}$
\STATE Initial LLM planner $f_{\text{LLM}}$ with parameters $\theta_{\text{LLM}}$
\STATE Initial RL policy $\pi_{\theta}$ with parameters $\theta_{\text{RL}}$
\STATE Mastery threshold $\tau$, number of cycles $C$

\ENSURE
\STATE Optimized planner $f_{\text{LLM}}^*$
\STATE Trained policy $\pi_{\theta}^*$

\STATE
\STATE Initialize subtask bank $\mathcal{B} \gets \emptyset$
\STATE Initialize candidate plans $\mathcal{P} \gets \{\}$
\STATE Initialize mastered subtasks $\mathcal{M} \gets \emptyset$

\STATE 
\STATE \textbf{Initial Plan Generation (Cycle 1):}
\STATE Extract subtasks: $\mathcal{S} \gets f_{\text{LLM}}(\mathcal{D}_{\text{train}})$
\STATE Build ontology: $\mathcal{O} \gets \text{BuildOntology}(\mathcal{S}, f_{\text{LLM}})$
\STATE Generate initial plans: $\mathcal{P_\text{initial}} \gets \text{ExpandPlans}(\mathcal{D}_{\text{train}}, \mathcal{O})$
\STATE
\STATE Fine-tune $f_{\text{LLM}}$ on $\mathcal{P_\text{initial}}$ using SFT
\STATE Generate training plans: $\mathcal{P} \gets f_{\text{LLM}}(\mathcal{D}_{\text{train}})$

\STATE 
\FOR{cycle $c = 1$ \TO $C$}
    
    \STATE 
    \STATE \textbf{Policy Training with Skill Curriculum:}
    \STATE Train $\pi_{\theta}$ on $\mathcal{P}$ using PPO with Skill Curriculum Learning
    \STATE Update mastered subtasks $\mathcal{M}$ based on success rates
    
    \STATE 
    \STATE \textbf{Plan Validation:}
    \STATE Execute $\pi_{\theta}$ with plans $P$ for multiple seeds
    \STATE For every plan $P$ in compute average success rate $SR(p)$
    \STATE Construct preference dataset $D_{\text{pref}}$
    
    \STATE 
    \STATE \textbf{LLM Fine-tuning:}
    \STATE Fine-tune $f_{\text{LLM}}$ on $\mathcal{D}_{\text{pref}}$ using DPO

    \STATE 
    \STATE \textbf{Plan Generation:}
    \STATE Select plans for new training epoch: 
    \STATE $\mathcal{P} \gets \text{SelectPlans} ( f_{\text{LLM}}, \mathcal{D}_{\text{train}}, \mathcal{P})$  
\ENDFOR

\STATE 
\STATE \textbf{return} $f_{\text{LLM}}, \pi_{\theta}$
\end{algorithmic}
\end{algorithm}

\section{DPO plans reprioritization}
 \label{app:reprioritization}
 
\begin{figure}[H]
    \centering
    \includegraphics[width=1\linewidth]{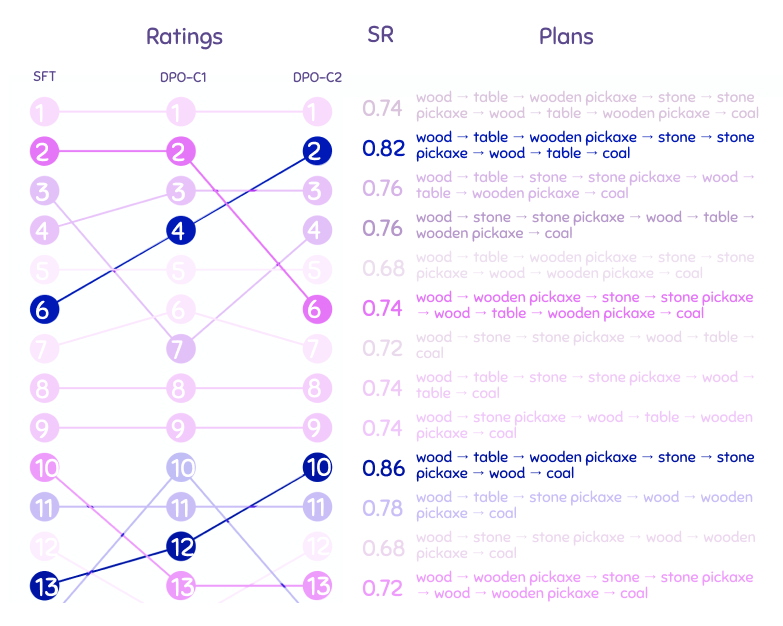}
    \caption{Example of DPO plan reprioritization for the instruction:\textit{"Forge a stone pickaxe and mine coal"}}
    \label{fig:dpo_reparam}
   
\end{figure}

% \begin{wrapfigure}{l}{0.65\textwidth}
% \vspace{-20px}
%   \includegraphics[width=0.65\textwidth]{figures/grouding_example.pdf}
%   \caption{Example of DPO plan reprioritization for the instruction:\textit{"Forge a stone pickaxe and mine coal"}}
%   \label{fig:dpo_reparam}
% \vspace{-20px}
% \end{wrapfigure}

\begin{table*}[t]
\small
\centering
\caption{Comparison of computational cost and performance across methods.}
\label{tab:compute_summary}
\begin{tabular}{lcccc}
\toprule
Method & RL Training (5B steps) & LLM Costs (2 cycles) & Total Cost & Atomic SR \\
\midrule
PPO-T (Baseline) & $\sim$120h & 0h & $\sim$120h & 0.03 \\
FiLM (Baseline) & $\sim$120h & 0h & $\sim$120h & 0.06 \\
SuperIgor / SI-DPO & $\sim$120h & $\sim$32h & $\sim$152h & $0.44 \pm 0.02$ \\
\bottomrule
\end{tabular}
\end{table*}

\section{Training Details: Policy Optimization}
\label{app:ppo_training}

Our low-level policy, which is responsible for executing individual subtasks, is trained using Proximal Policy Optimization (PPO). The agent's goal at this stage is to learn an optimal strategy for completing a given subtask based on its visual observations. The standard clipped surrogate objective for PPO is defined as:

{\small
\begin{equation*}
\mathcal{L}^{\text{PPO}}(\theta) = \mathbb{E}_t \left[ \min ( \rho_t(\theta) \hat{A}_t, \text{clip}(\rho_t(\theta), 1 - \epsilon, 1 + \epsilon)\hat{A}_t) \right],
\end{equation*}
}

where \( \rho_t(\theta) = \frac{\pi_\theta(a_t \mid o_t)}{\pi_{\theta_{\text{old}}}(a_t \mid o_t)} \) is the probability ratio and \( \hat{A}_t \) is the estimated advantage at timestep \( t \).

% Про базовую архитектуру

Our agent's policy and value functions are parameterized by a single neural network with a shared multimodal feature extractor and separate actor and critic heads. The visual stream processes the \( 63 \times 63 \) pixel image with \( 3 \) channels observations using a three-layer Convolutional Neural Network (CNN). Each convolutional layer utilizes 32 filters with a $5 \times 5$ kernel, followed by a ReLU activation and max-pooling. For the language stream, textual instructions are encoded using a pre-trained BERT model (\texttt{bert-base-uncased}), and we use the embedding of the \texttt{[CLS]} token as the final text representation.

The flattened output of the CNN and the text embedding are then concatenated to form a unified multimodal representation. This combined feature vector is fed into two separate feed-forward networks: the \textbf{actor head}, which outputs the logits for the categorical action distribution, and the \textbf{critic head}, which outputs a scalar estimate of the state-value function.

\section{Training Details: LLM Fine-Tuning }
\label{app:llm_finetuning}

To improve the high-level planner (the LLM), we employ a reinforcement learning-based feedback loop. The planner generates a sequence of subtasks (a plan), which is then executed by the PPO agent. The final outcome of the agent's execution (e.g., task success or failure, efficiency) serves as a signal to update the planner.

\textbf{Direct Preference Optimization (DPO).}  
This method aligns the model toward preferred completions using pairwise preference data. The DPO loss is:

{\footnotesize
\begin{equation*}
\mathcal{L}^{\text{DPO}} = -\log \sigma\left( \beta \left( \log \pi(x^{+} \mid q) - \log \pi(x^{-} \mid q) \right) \right),
\end{equation*}
}

where \(x^{+}\) and \(x^{-}\) are preferred and less preferred plans for instruction \(q\), and \(\beta\) is a temperature parameter.

\section{Agent’s plan following}
 \label{app:plan_following}
Figure \ref{fig:dpo_reparam} presents the example of how the agent follows the plan and chooses actions. For each subtask, there are two frames: the first shows the observation up to the moment when the agent takes the action that completes the subtask, along with the action distribution at that time; the second shows a few timesteps later, when the agent decides to skip the subtask in order to solve a new one.

\begin{figure*}[!ht]
    \vspace{-10px}
    \centering
    \includegraphics[width=0.9\linewidth]{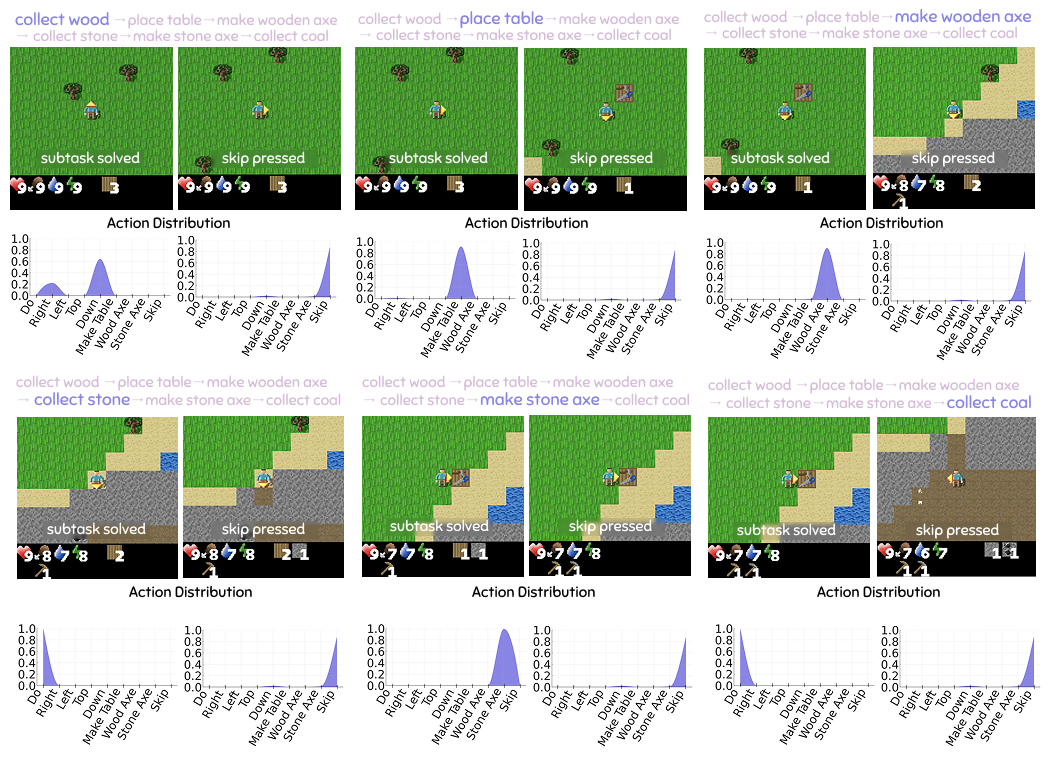}
    \caption{Example of how the agent follows the plan and chooses actions.}
    \label{fig:episode_example}
     \vspace{-10px}
\end{figure*}

\section{Compute Resources}
\label{app:compute_resources}

All experiments were conducted on a high-performance computing cluster equipped with nodes containing 1 NVIDIA A100 GPU with 80 GB of VRAM. Each node was powered by an 12 CPU Cores CPU with 96 GB of system RAM.

The total computational budget can be broken down into two primary stages:

\textbf{Policy Training and Evaluation.} The primary computational cost stems from training the PPO agent. Each full training run for a single configuration up to 5 billion environment steps took approximately 120 GPU-hours for all baselines including PPO-T, PPO-T+, FiLM and SI executor training. Reproducing all presented experiments, including the baseline comparisons and ablation studies, required a total of 10 such training runs.

\textbf{LLM Training and Generation.} The initial generation of plans using the Qwen2.5-14B-Instruct model for the entire dataset required approximately 1-2 GPU-hours on a single NVIDIA A100 GPU. Epoch of finetuning LLM with DPO on evaluated plans takes approximately 15 GPU-hours.

As shown in Table~\ref{tab:compute_summary}, incorporating LLM-based planning increases the total computational cost moderately, while yielding a substantial improvement in task success rate.

% \section{Limitation}
% \label{app:limitation}

% Although our proposed method demonstrates a promising direction for integrating large language models with reinforcement learning for instruction-following tasks, it is not without limitations.

% A primary limitation is the ambiguity in the attribution of failures. When the RL agent fails to complete a given plan, it is difficult to determine whether the failure stems from a flawed plan generated by the LLM or from inadequately trained policy in the RL agent. This ambiguity complicates the fine-tuning process for the language model, as the feedback signal may incorrectly penalize a viable plan that the agent was simply unable to execute. This can lead to a noisy training signal and potentially degrade the LLM's planning capabilities.

\section{Training Details: Hyperparameters}
\label{app:hyperparams}

The hyperparameters for our experiments are detailed in Table \ref{tab:hyperparams}. For the PPO agent training, we adopt the configuration from the original CrafText baseline study \citep{volovikova2025craftext}. For the LLM planner fine-tuning, we use a Q-LoRA approach with a comprehensive set of parameters optimized for efficient large model training.

\begin{table}[H]
\centering
\scriptsize
\caption{Hyperparameters used for training the low-level agent and fine-tuning the high-level planner.}
\label{tab:hyperparams}
\begin{tabular}{lc}
\toprule
\textbf{Hyperparameter} & \textbf{Value} \\
\midrule
\multicolumn{2}{c}{\textit{PPO Agent Training}} \\
\midrule
Learning rate & 0.0002 \\
Discount factor ($\gamma$) & 0.99 \\
GAE lambda ($\lambda$) & 0.95 \\
Clipping epsilon ($\epsilon$) & 0.2 \\
PPO epochs & 4 \\
Number of minibatches & 8 \\
Entropy coefficient & 0.01 \\
Value function coef. & 0.5 \\
Activation function & Tanh \\
Hidden layer size & 512 \\
\midrule
\multicolumn{2}{c}{\textit{LLM Planner Fine-Tuning}} \\
\midrule
Base model & Qwen2.5-14B-Instruct \\
Training epochs & 1 \\
Learning rate (SFT) & 2e-4 \\
Learning rate (DPO) & 1e-5 \\
Beta (DPO) & 0.5 \\
Optimizer & Paged AdamW (32-bit) \\
LR scheduler & Cosine \\
Warmup ratio & 0.03 \\
Batch size (per device) & 16 \\
Gradient accumulation & 1 \\
Gradient clipping norm & 0.3 \\
Weight decay & 0.001 \\
Mixed precision & bf16 \\
\midrule
\multicolumn{2}{c}{\textit{LoRA Configuration}} \\
\midrule
LoRA rank (r) & 64 \\
LoRA alpha ($\alpha$) & 16 \\
LoRA dropout & 0.1 \\
\midrule
\multicolumn{2}{c}{\textit{Quantization (4-bit)}} \\
\midrule
Quantization type & nf4 \\
Compute dtype & float16 \\
\bottomrule
\end{tabular}
\end{table}

% % TODO: нужно ли правда добавлять ахах (???)
% \section{Agent’s plan following}
%  \label{app:plan_following}
 
% \begin{figure*}[!ht]
%     % \vspace{-10px}
%     \centering
%     \includegraphics[width=0.9\linewidth]{figures/following_example.png}
%     \caption{Example of how the agent follows the plan and chooses actions.}
%     \label{fig:dpo_reparam}
% \end{figure*}

%\clearpage

\begin{figure*}[t!]

\section{AI Assistants Usage}
AI assistants were used solely for language editing and text refinement. All scientific contributions and conclusions are the sole responsibility of the authors.

\section{Plans Generation: Prompt for plan generation}
\label{app:zoe_prompts}

\vspace{4mm}

%\subsection*{Main Prompt Template}

\vspace{4mm}

\scriptsize
\begin{verbatim}
You control an agent in a 2D game with simplified Minecraft environment. 
You will need to provide a detailed step-by-step plan for following the user's instructions. 
You must include all the preliminary steps that it needs to complete.

You are controlling an agent in a 2D game set within a simplified Minecraft-like environment. 
The agent starts from scratch with an empty inventory and no gathered resources. 
Your task is to generate a step-by-step plan that enables the agent to follow a given user instruction.

What you must do:
- Break down the instruction into atomic actions the agent needs to perform.
- Include all necessary preliminary steps, such as gathering or crafting resources.
- Assume the agent has nothing at the beginning — you must plan from the ground up.
- Output your answer as a Python list of strings.
- Each string must represent one atomic skill invocation, written on a separate line.

Format for each step:
"skill_name(arg1 = value1, arg2 = value2, ...)"
- skill_name: the name of the primitive skill or action the agent will execute.
- Inside the parentheses, list all required arguments with their names and corresponding values.

Example:
gather_resource(resource_type = wood)

Each of the step agents will be implemented without knowledge of what it did before, 
so it can only rely on observation and the current step. 
Therefore, each step must be self-sufficient and not require knowledge of past steps.

"If the instruction doesn't specify what the agent needs to do and is more general—like 
'Explore the world' or 'Go out and examine the world around you'—send explore(object=world). 
In this case, the plan should consist of only one step: "explore(object=world)"."

Send your answer as a python list.
Instruction: Make a pickaxe from wood
Answer: 
["gather_resource(resource_type = wood)",
"gather_resource(resource_type = wood)",
"create_item(item_type = table)", 
"gather_resource(resource_type = wood)", 
"gather_resource(resource_type = wood)", 
"create_item(item_type = wooden_pickaxe)"]

Send your answer as a python list.
Instruction: $INSTRUCTION$  
Answer:
\end{verbatim}

%\vspace{25px}
 \end{figure*}

% \end{document}

\end{document}